\documentclass[]{article}

\usepackage{graphicx} 
\usepackage{bm}
\usepackage{xcolor}
\usepackage{longtable}
\usepackage{supertabular}
\usepackage[utf8x]{inputenc}
\usepackage{amssymb}

\begin{document}

\title{Trajectory saliency detection using consistency-\\oriented
  latent codes from a recurrent auto-encoder}

\author{Léo~Maczyta,
        Patrick~Bouthemy\thanks{L. Maczyta and P. Bouthemy are with Inria, Centre
  Rennes - Bretagne Atlantique, Campus de Beaulieu, Rennes,
  35042, France.},
        and~Olivier~Le~Meur
\thanks{O. Le Meur is with Univ Rennes, CNRS, IRISA Rennes, Campus de
  Beaulieu, Rennes, 35042, France.}
}

\section*{IEEE copyright notice}

\textcopyright~2021 IEEE. Personal use of this material is
permitted. Permission from IEEE must be obtained for all other uses,
in any current or future media, including reprinting/republishing this
material for advertising or promotional purposes, creating new
collective works, for resale or redistribution to servers or lists, or
reuse of any copyrighted component of this work in other works.\\
IEEE Transactions on Circuits and Systems for Video Technology \\
DOI: 10.1109/TCSVT.2021.3078804

\newpage

\maketitle

\begin{abstract}

  In this paper, we are concerned with the detection of progressive
  dynamic saliency from video sequences. More precisely, we are
  interested in saliency related to motion and likely to appear
  progressively over time. It can be relevant to trigger alarms, to
  dedicate additional processing or to detect specific
  events. Trajectories represent the best way to support progressive
  dynamic saliency detection. Accordingly, we will talk about
  trajectory saliency. A trajectory will be qualified as salient if it
  deviates from normal trajectories that share a common motion pattern
  related to a given context. First, we need a compact while
  discriminative representation of trajectories. We adopt a (nearly)
  unsupervised learning-based approach. The latent code estimated by a
  recurrent auto-encoder provides the desired representation. In
  addition, we enforce consistency for normal (similar) trajectories
  through the auto-encoder loss function. The distance of the
  trajectory code to a prototype code accounting for normality is the
  means to detect salient trajectories. We validate our trajectory
  saliency detection method on synthetic and real trajectory datasets,
  and highlight the contributions of its different components. We show
  that our method outperforms existing methods on several scenarios
  drawn from the publicly available dataset of pedestrian trajectories
  acquired in a railway station \cite{Alahi2014}.

\end{abstract}

\section{Introduction}

Saliency estimation in images is a common task. It
may operate as a pre-attention mechanism or be a goal in itself. Its
output usually consists of saliency maps providing the probability of each pixel to be salient. Actually, saliency estimation is close to visual attention, object detection, or even image segmentation. This paper is concerned with the detection of salient trajectories, which is a particular but important kind of saliency in videos. In what follows, we first draw up a broad perspective of saliency in images to properly position this type of saliency.

There is not really a single formal definition of image saliency. It is something that stands out from its surroundings in the image and
attracts attention. In static images, this generally refers to one (or a few) prominent object(s) in the foreground as adopted in most
challenges and benchmark datasets.  It becomes a prominent moving
object in the foreground when it comes to videos, and one talks about dynamic saliency.

In contrast, we will only take into account \emph{motion} to define
dynamic saliency, that is, local motion that deviates from the main
motion of its surrounding. It may seem more specific, since it does
not exhibit any notion of object. However, it can also be considered
more general. Indeed, it involves all the cases where the only
differentiating factor is movement, whatever the appearance is. A
typical example is an individual walking against a crowd. It can occur in many diverse applications, such as monitoring of urban and road traffic \cite{Roy2019, Barata2019}, safety of public areas prone to dense crowd \cite{Perez2017}, study of cell dynamics in bio-imaging \cite{Roudot2017}, or identification of adverse weather conditions in meteorological image sequences \cite{Papin2000}. Motion saliency detection can be useful to trigger alarms, to focus additional processing or to allow for the detection of a specific event.

Here, we are going one step further, since we are interested in
detecting \emph{progressive} dynamic saliency. In other words, we are talking about saliency that may not be visible instantaneously, but rather progressively appears over time. Trajectories are the most natural features to support the detection of progressive dynamic saliency. Accordingly, we will talk about \emph{trajectory saliency}. Indeed, trajectory-based detection of unexpected events has emerged as a growing need in many applications, possibly under
different names
\cite{Roy2019,Mousavi2015,Fan2018,Laxhammar2014,Piciarelli2008}.

Furthermore, salient trajectories and anomalous (or abnormal) trajectories are slightly different notions. In general, abnormal trajectories are abnormal in themselves relatively to an expected behavior in a given application. In our view, salient trajectories are salient relative to a majority in a set of trajectories or to surrounding trajectories. Thus, its state depends on the present context; a trajectory may be salient in a given context but not salient in another one, whereas trajectory anomaly is an intrinsic state. By the way, we will use the term “normal trajectories” for the majority or surrounding trajectories for convenience. A salient trajectory by definition attracts attention, and this, whatever its real nature (simply different, singular, unusual, inappropriate, potentially dangerous...). Besides, an abnormal trajectory is a case of salient trajectory, but not the other way around.

To get an idea, let us give the following concrete example in a real-life scenario. A person walking against a crowd will produce a trajectory that is salient with respect to the trajectories of the crowd. However, its trajectory is not intrinsically abnormal. The same person walking in the same way, but this time within a crowd, will not produce a salient trajectory. On the other hand, a person staggering in the crowd will produce an abnormal trajectory regardless of the overall movement of the crowd.

Two main issues then arise: which representation for the
trajectories and which method for trajectory saliency detection.  We
aim to design a method as general and efficient as possible. To do
this, learning-based approaches appear nowadays superior to
handcrafted representations. If enough data are available, we can
learn the trajectory representation with neural networks. We will resort to Recurrent Neural Network (RNN), as done for example in \cite{Ma2018,Yao2017}. More specifically, we will adopt a LSTM-based auto-encoder. Besides, we will design a decision algorithm to solve the second issue.

Our main contributions can be summarised as follows: 1) Characterization of the notion of progressive dynamic saliency in videos and the associated trajectory saliency paradigm; 2) Introduction of a consistency constraint in the training loss of the LSTM-based auto-encoder producing the latent representation of the trajectories; 3) Trajectory-saliency detection designed in the latent trajectory-representation space; 4) Thorough and comparative evaluation on a large dataset of real pedestrian trajectories.

The rest of the paper is organised as follows. Section~\ref{sec:rel}
is concerned with related work. Section~\ref{sec:traj_code} presents
our latent trajectory representation based on a LSTM
auto-encoder. Section~\ref{sec:sal_est} deals with the detection of
trajectory saliency. In Section~\ref{sec:datasets}, we present the two trajectories datasets, a synthetic one and a real one. In addition, we design two evaluation sets, issued from the real dataset, for trajectory-saliency detection. In Section~\ref{sec:res}, we report experimental results on the synthetic and real trajectory datasets, including an objective comparative evaluation, and additional investigations on the main stages of the method. Section~\ref{sec:concl} contains concluding comments.

\section{Related work}
\label{sec:rel}

Dynamic saliency estimation in videos consists in identifying moving objects that depart from their context. We can classify the existing
approaches in two categories. The first category assumes that salient
elements are characterised by distinctive appearance and/or {
motion \cite{Karimi2016,Le2016,Mancas2011,Kim2014,Wang2015}. }
Recent methods leverage deep learning techniques,
either by estimating video saliency end-to-end \cite{Wang2018}, or by
first extracting intermediate deep features \cite{Le2018}. A second
category of video saliency methods aims to predict which areas of the
image attract the gaze of human observers. The objective is then to
leverage eye-tracking data in order to predict efficiently
eye-fixation maps \cite{Chaabouni2016, Liu2014, Qiu2018}.

Although significant progress has been done during the last decade,
there are still many open issues to address. Benchmarks used for video
saliency evaluation typically consider only one or a few moving
objects in the foreground of the viewed scene \cite{Ochs2014,Perazzi2016}. For such configuration, dynamic saliency can be
directly estimated from a couple of successive images and possibly
from optical flow. However, for more complex situations, this approach
may become less adequate.
For a denser configuration of independently moving objects, and longer periods of observation, a change of paradigm can lead to better insights. This can be achieved by focusing on trajectories instead of raw visual content, as done in \cite{Huang2014} to estimate saliency maps in videos.

Trajectories can be obtained through several means. They can be
directly provided with sensors such as GPS \cite{Endo2016}, which have
become ubiquitous and can generate large amount of data. For natural
videos, a tracker is usually applied to a unique camera as in
\cite{Roy2018}, but more complex settings can be considered. A network
of cameras with a processing pipeline including tracking and
re-identification can provide trajectories of people over a large area
\cite{Alahi2014}. In specific contexts such as bio-imaging, trackers
taking into account the expected properties of image content can be
developed \cite{Chenouard2013}.

When dealing with trajectories, extracting a compact representation is
useful for many tasks. A trajectory auto-encoder network can be an
efficient and general way to obtain such a representation without
making any prior assumption \cite{Yao2017, Chow2018, Co-Reyes2018, Su2016}. In \cite{Yao2017}, this representation is obtained with a
recurrent auto-encoder and is exploited for trajectory clustering. In the following, we will refer to this method as TCDRL that stands for Trajectory Clustering via Deep Representation Learning, which is the title of the paper. In \cite{Su2016}, the representation is leveraged for crowd scene understanding. In \cite{Co-Reyes2018}, the authors use a latent representation in a reinforcement-learning framework. The authors of \cite{Chow2018} take into account spatial constraint of the environment, such as walls, for the estimation of the trajectory representation, thanks to the optimisation of a spatially-aware objective function.

With trajectory data, estimating dynamic saliency can be achieved by
searching for anomalous patterns. In   \cite{Piciarelli2008}, the authors make use of single-class SVM clustering. In \cite{Roy2018}, the authors
consider road user trajectories. They design the DAE method that
reconstructs trajectories with a convolutional auto-encoder. Then,
they assume that poorly reconstructed trajectories are necessarily
abnormal, since the latter are likely to be rare or even
missing in the training data. In an extended work presenting the ALREC
method \cite{Roy2019}, the same authors adopt a Generative Adversarial
Network (GAN) to classify trajectories as being normal or
salient. They train the discriminator to distinguish between normal and
abnormal trajectories from reconstruction errors. While the threshold between normality and abnormality is automatically set with the GAN, the reconstruction error remains the core principle of the method. In \cite{Ma2018} that describes the AET method, the authors train a recurrent auto-encoder for each normal trajectory of a training set, and compute the self-reconstruction error for every one. For each trained network, they compare the reconstruction error obtained for a tested trajectory to the corresponding self-reconstruction error. Accordingly, the tested trajectory is classified as abnormal or not. In \cite{Ji2020}, the authors leverage a LSTM-based recurrent network to estimate the representative of each cluster of trajectories, and a distance between the tested trajectory and each cluster representative is used to decide whether the tested trajectory fits one group or none.

The trajectory representation may not be limited to moving objects, as
it can be relevant for time series data in general. In \cite{Fan2018},
the authors consider building energy data over time. They follow
similar principles as for other types of trajectories, and base their
anomaly detection method on the reconstruction error provided by an
auto-encoder. { In \cite{Wu2017}, the authors consider that trajectories are paths in a graph. They also adopt recurrent networks for this particular kind of trajectories.

The paradigm of inferring trajectory abnormality from the reconstruction error of an auto-encoder has some limitations. First, it assumes that the learned models will generalise badly to unseen data. In addition, abnormality is handled in an absolute way. An element is stated as abnormal or not in itself. Therefore, it does not immediately extend to saliency. Indeed, as far as saliency is concerned, the same element can be salient or not depending on its context. This key issue needs to be addressed explicitly.} We will introduce an original framework to properly deal with that.

Recently, significant progress has been made in reducing the
supervision burden when training. An unsupervised stage can provide a
strong pre-trained network for fine-tuning. In \cite{Wang2019},
tracking is performed with an unsupervised framework.  For the
trajectory prediction task \cite{Amirian2019}, the GAN framework is a
mean to train a model without requiring manual annotations. Taking
inspiration from these works, we will similarly seek to reduce as much
as possible the annotation cost to ensure a broad applicability.

\section{Latent trajectory representation}
\label{sec:traj_code}

\begin{figure*}[!htbp]
  \centering
  \includegraphics[width=\linewidth]{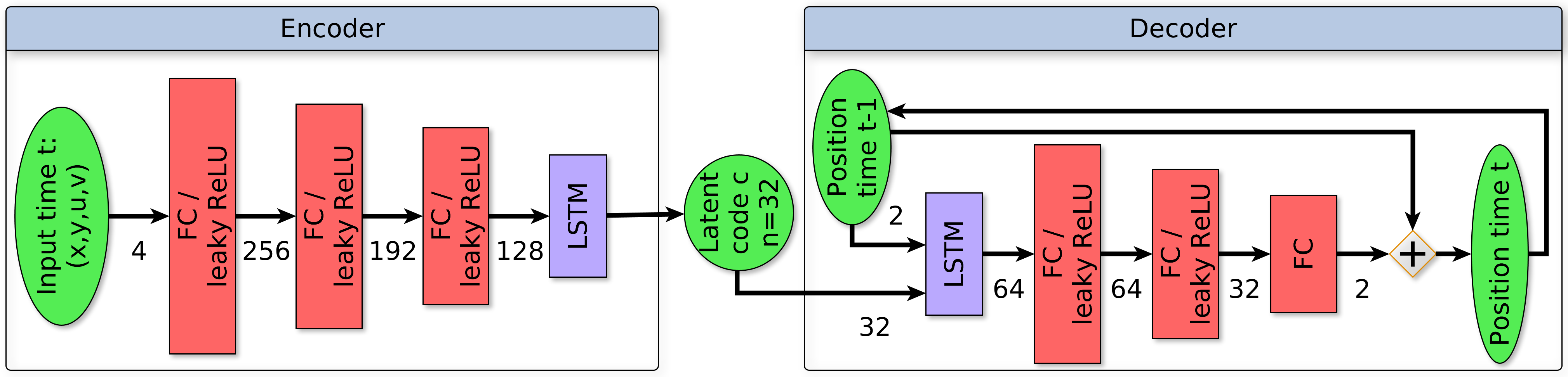}
  \caption{Recurrent auto-encoder network computing the latent
    representation of trajectories. Numbers indicate the input
    dimension for each layer.}
  \label{fig:net}
\end{figure*}

Our objective is to detect salient trajectories within a set of
trajectories, the vast majority of which are normal trajectories. In
the following, a scenario $\mathcal{S}$ will consist of a set of
normal trajectories for a given context and possibly a few salient
trajectories with respect to the same context. Indeed, a trajectory is
not salient in itself, but with respect to a given context. Salient
trajectories are supposed to depart from the motion pattern shared by
the normal trajectories.

We need a relevant representation of the trajectories. It has to be
compact to form the basis of an efficient classification method, while
descriptive enough to distinguish normal and salient trajectories. It
is preferable to transform trajectories of any length into a
constant-length representation, in order to enable adaptability to any
scenario. It should be applicable to both 2D and 3D trajectories. In
that vein, hand-crafted representations based on sophisticated
mathematical developments were investigated in the past, as in
\cite{Hervieu2008}. The latter provided representations invariant to a large class of geometrical transformations, including translation,
rotation, scale, and able to achieve fine classification.

However, it is now more tractable to learn the targeted representation with the advent of neural networks, providing enough data are available for training \cite{Yao2017,Su2016}. An appropriate means is the use of auto-encoders. The internal (latent) code in the auto-encoder will provide us with the desired representation. Besides, a trajectory exhibits a sequential nature inherent to the time dimension, making a recurrent network a suitable choice. Below, we will first describe the representation we have designed for trajectories based on these requirements. Then, in
Section~\ref{sec:sal_est} we will explain how we leverage it to detect salient trajectories.

\subsection{LSTM-based network for trajectory representation}
\label{sec:enc_dec}

Our approach is to compute the trajectory representation with a
recurrent auto-encoder network. More specifically, we adopt a Long
Short-Term Memory (LSTM) network.  The objective of an auto-encoder is
to reconstruct its input, by relying on an intermediate compact
representation, acting as the code of the input \cite{Ribeiro2018}. An
auto-encoder has the advantage of being unsupervised, since no manual
annotation is required for training.

This architecture is inspired from existing networks, such as the
generator of \cite{Amirian2019}. We similarly build a stack of fully
connected layers to process the data at each time instant, but we
dedicate the temporal processing to the recurrent LSTM units. This
choice allows us to easily handle trajectories of variable length, in
contrast to 1D temporal convolutions.

\begin{figure}[!htbp]
  \centering
  \includegraphics[width=\linewidth]{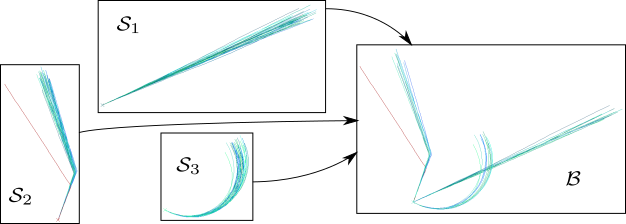}
  \caption{Trajectories of three scenarios $\mathcal{S}_1$,
    $\mathcal{S}_2$, $\mathcal{S}_3$ (from the STMS dataset described
    in Section~\ref{sec:STMS}) are displayed on the left. A batch
    $\mathcal{B}$ composed of trajectories sampled from the three
    scenarios is represented on the right.}
  \label{fig:batch_traj}
\end{figure}

The architecture of the auto-encoder supplying the trajectory
representation is summarised in Figure~\ref{fig:net}. The input of the network at time $t$ is composed of the concatenation of the 2D
position $(x_t, y_t)$ and velocity $(u_t, v_t)$. For the processing of the whole trajectory $\mathcal{T}_i$, it gives a total of $4\tau$
features if $\tau$ is the length of the trajectory, i.e., the number
of points it contains. In practice, we assimilate the velocity to the displacement. Although the displacement can be deduced from two
successive positions, it still helps the network properly estimate the trajectory representation at a negligible cost.

The 2D-space can be the image plane and the velocity vector given by
the optical flow. If the 3D movements are planar, the 2D-space could
be the ground plane of the 3D scene and trajectories (positions and
velocities) are referred to this plane. An extension to full 3D
trajectories would be straightforward.

The exact number of layers and their dimensions have been set during
preliminary experiments. In particular, we set the dimension $n$ of
the code $c$ to $32$, as this value allows us to store enough
information to represent the motion patterns, while being shorter that the typical trajectory representation of length~$4\tau$. In the
experiments, we consider trajectories involving a number of positions between 20 to 200.

To process one time step, the encoder activates a succession of three
fully connected layers with output of dimension 256, 192 and 128
respectively, with the leaky ReLU activation function. { The fully connected layers are followed by a single LSTM layer whose output is of
dimension $n=32$. The output is precisely the hidden state and provides the latent code $c_i \in\mathcal{R}^{n}$ representing the whole trajectory $\mathcal{T}_i$.

The decoder reconstructs the trajectory from code $c_i$, obtained once trajectory $\mathcal{T}_i$ has been fully processed by the encoder. A LSTM-based decoder, involving one single recurrent layer, takes as input the code $c_i$ concatenated with the
position at time $t-1$, and produces a vector of 64 components}. The
decoder is only expected to expand the information compressed in the
code. The predicted position for the current time instant is obtained through two fully connected layers, followed by the leaky ReLU non linearity, whose respective output dimensions are 64 and 32, and one final fully-connected layer of output dimension 2 representing the displacement. The position is given by the sum of the displacement and of the previous position. Indeed, preliminary experiments showed that predicting the displacement is more robust than predicting the position directly. Let us mention that, for the reconstruction of a trajectory, we assume that its length is available (in practice, we compute it from the list of positions). The number of parameters for the whole network is 127,970.

To train the auto-encoder, we adopt the reconstruction loss
$\mathcal{L}_r$, defined as follows:

\begin{equation}
  \label{eq:loss_rec}
  \mathcal{L}_r = \sum_{\mathcal{T}_i \in \mathcal{B}} \sum_{t=t_{init}}^{t_{final}} (x^i_t - \hat{x}^i_t)^2 + (y^i_t - \hat{y}^i_t)^2,
\end{equation}
with $t_{init}$ and $t_{final}$ the initial and final time instants of the trajectory $\mathcal{T}_i$, $(x^i_t, y^i_t)$ its position at time $t$, $(\hat{x}^i_t, \hat{y}^i_t)$ the predicted position and
$\mathcal{B}$ a batch of trajectories. A batch can be composed of
trajectories taken from several scenarios as illustrated in
Figure~\ref{fig:batch_traj}. It is a standard practice to train
auto-encoders with the reconstruction error as loss function
\cite{Yao2017, Ribeiro2018}. Since the positions entirely define the
trajectory, the loss $\mathcal{L}_r$ involves only positions.

Once trained, the auto-encoder is ready to provide a code representing the whole trajectory at test time. However, there is so far no guarantee that two similar trajectories will be represented with close codes. This is an important issue since we will exploit the codes to predict saliency.  It motivates us to introduce a second loss term for code estimation.

\subsection{Consistency constraint}

\begin{figure}[!htbp]
  \centering
  \includegraphics[width=\linewidth]{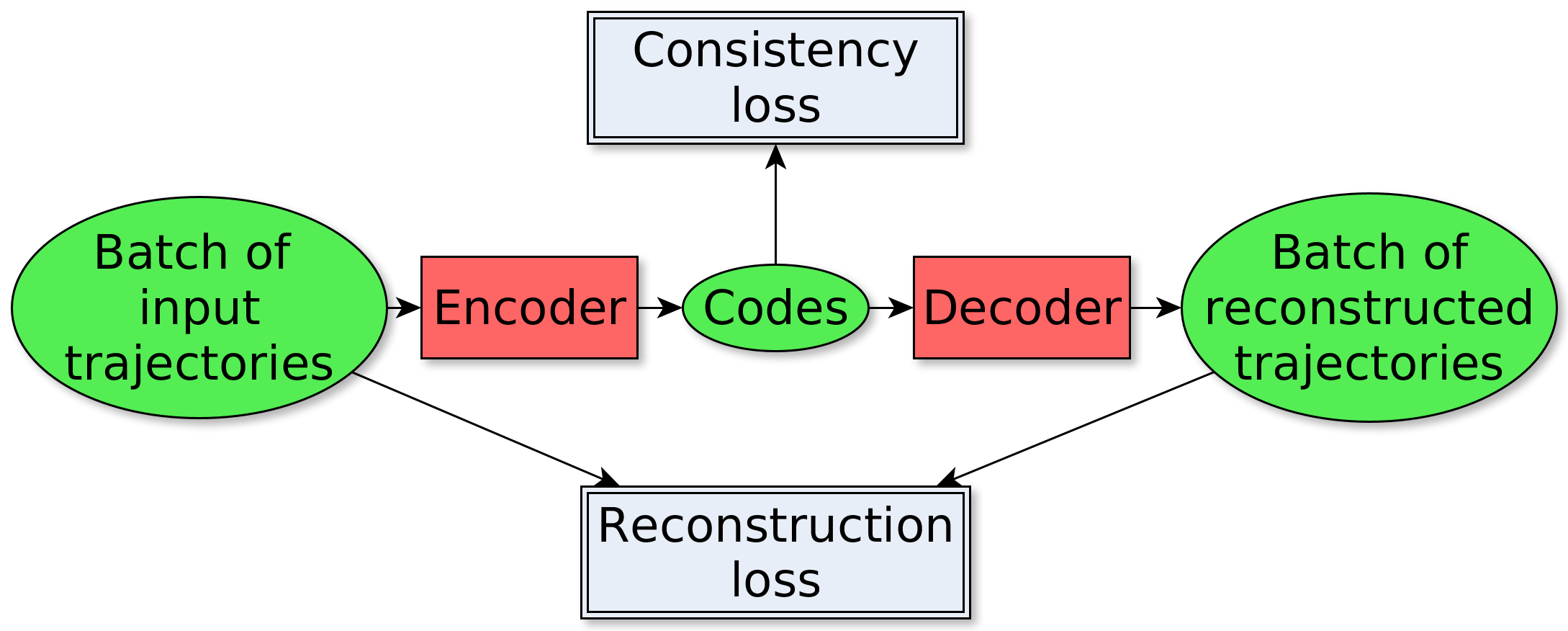}
  \caption{Training scheme of the recurrent auto-encoder with the two
    associated losses.}
  \label{fig:glb}
\end{figure}

Our objective is to represent similar trajectories with close codes
and dissimilar trajectories with distant codes to make saliency
prediction easier. We somehow take inspiration from the paradigm of
deep metric learning \cite{Schroff2015}. We want to ensure the best
possible consistency in the representation of normal trajectories,
while drastically minimising, or even discarding, manual
annotation. To this end, we complement our auto-encoder network with a consistency constraint, acting as a second loss term. Figure~\ref{fig:glb} summarises our overall training scheme.

Let us recall that we address the problem of detecting salient
trajectories that are supposed to be rare within a group of normal
trajectories. The normal trajectories of a given scenario are supposed to all follow a consistent motion pattern. We then define the following consistency loss term $\mathcal{L}_c$ applied to each
scenario~$\mathcal{S}_k$ present in the batch $\mathcal{B}$:
\begin{equation}
  \label{eq:const_loss}
  \mathcal{L}_c = \sum_{\mathcal{S}_k \triangleright  \mathcal{B} } \; \sum_{\mathcal{T}_i \in \mathcal{S}_k \cap \mathcal{B}  }  ||c_i - \widetilde{c}_k||_2,
\end{equation}
with $c_i$ the code of the trajectory $\mathcal{T}_i$ estimated by the network, and $\widetilde{c}_k$ the median code for the trajectories of the batch issued from the scenario $\mathcal{S}_k$. The $\triangleright$ operator designates scenarios that contribute to the batch.

Applying the consistency loss to all the trajectories of the scenario means obviously that salient trajectories might be involved as well if there are any. On one hand, this setting may seem contradictory to our objective.  However, let us point out again that salient trajectories are rare by definition. Therefore, they are unlikely to affect the value of the median code. In addition, the use of the L2 norm instead of the square in the consistency loss function~(\ref{eq:const_loss}) further limits the impact of the consistency constraint on the salient trajectories.

The full loss will add the quadratic loss term $\mathcal{L}_r$ defined in eq.(\ref{eq:loss_rec}) to the consistency constraint of
eq.(\ref{eq:const_loss}).  One of the roles of $\mathcal{L}_r$ is to mitigate the effect of $\mathcal{L}_c$ on salient
trajectories. Indeed, the combination of the two terms will express
the trade-off between the reconstruction accuracy and the code
distinctness for the salient trajectories. The large imbalance between salient and normal elements is not an issue, unlike for supervised classification. It even becomes an advantage for the correct training of our network by limiting the possible impact of salient trajectories on the consistency assumption. Besides, the addition of the consistency term still preserves an unsupervised training.

The final expression of the loss taking into account reconstruction
and consistency terms is given by:
\begin{equation}
  \label{eq:unsup_total_loss}
  \mathcal{L} = \mathcal{L}_r + \beta \mathcal{L}_c,
\end{equation}
where $\beta$ is a weighting parameter to balance the impact of the
two loss terms.
We have defined a method to estimate a latent code to represent
trajectories. It is specifically designed to obtain close codes for
similar trajectories and distant codes for potentially salient
trajectories.

\section{Trajectory saliency detection}
\label{sec:sal_est}

Once trained, the recurrent auto-encoder provides us with codes of
input trajectories. Now, we have to define an algorithm to detect
salient trajectories against normal ones. Let us recall that normal
trajectories share a common motion pattern with respect to a given
context. More specifically, we consider a scenario $\mathcal{S}$ of
the test dataset, containing normal and possibly salient trajectories,
$\mathcal{S} = \{\mathcal{T}_i\}$. The trajectories are represented by
their respective latent codes $\{c_i, c_i \in \mathcal{R}^n\}$. The
scenario $\mathcal{S}$ will be further defined when reporting each
experiment.

\subsection{Distance between codes}
\label{subsec:dis}

First, for comparison purpose, we need to characterise the normal
class by a "generic code". It will be used to classify trajectories into normal and salient ones. Due to the possible presence of salient trajectories in $\mathcal{S}$, the generic code will be given by the median of codes over $\mathcal{S}$, denoted by $\widetilde{c}$. Since we deal with a \textit{n}-component code, we compute the component-wise median code. Each component of $\widetilde{c}$ is the median value of the corresponding components of the codes $c_i$ over $\mathcal{S}$.

The classification of a trajectory $\mathcal{T}_i$, as normal or
salient, will leverage the distance $d_i$ between its code $c_i$ and
the median code $\widetilde{c}$~:
\begin{equation}
  \label{eq:d_i}
  d_i = || c_i - \widetilde{c} ||_2.
\end{equation}

The distances related to normal trajectories are expected to be
smaller than the ones related to salient trajectories. To make the
comparison invariant to the distance range, the empirical average of
the distances $d_i$, noted $\bar{d}$, and the standard deviation of
the $d_i$, noted $\sigma$, are computed over scenario
$\mathcal{S}$. We normalise the distance $d_i$ to get the following
descriptor $q_i$ for each trajectory:
\begin{equation}
  \label{eq:quo}
  q_i = \frac{|d_i - \bar{d}|}{\sigma}.
\end{equation}

With this definition, $q_i$ belongs to $\mathbb{R}^+$, with value
ideally close to zero for normal trajectories. The status of each
trajectory will be inferred from the descriptor $q_i$, as described
below.

\subsection{Inference of trajectory saliency from normalised distance}
\label{sec:sal_test}

Descriptors $q_i$ account for the possible deviation from the normal
motion of the scenarios for each trajectory
$\mathcal{T}_i$. Accordingly, a natural way to predict saliency is to assume that trajectories with a descriptor $q_i$ greater than a given threshold $\lambda$ are salient. We have investigated two ways of setting this threshold.

\subsubsection{\textit{p}-value method}

The \textit{p}-value scheme aims to statistically control the number
of false detections and enables to automatically set $\lambda$. We
assume that the $q_i$ descriptors for normal trajectories follow a
known probability distribution. We then need to estimate its
parameters to fit the empirical distribution of the $q_i$ descriptors
of a representative set of normal trajectories. In order to identify
candidate distributions, we looked at several histograms of $q_i$
descriptors and observed that their distribution is
skewed. Accordingly, the tested probability distributions were the
Weibull distribution \cite{Weibull1939}, the Dagum distribution
\cite{Dagum1977} in the standardised form and the Dagum distribution
in the general form. We selected the latter as explained in
Appendix~\ref{sec:alt_lambda}.

Once the three parameters of the general Dagum distribution are
estimated, we can fix the \textit{p}-value and then get the threshold
value $\lambda$. In an ideal case, the estimated distribution should
correspond also to the test data. At this point, let us emphasise that we have no guarantee that this is the case. Indeed, the $q_i$
descriptors depend on the presence of outliers, that is, salient
trajectories, through the normalisation with $\bar{d}$ and $\sigma$,
the computed mean and standard deviation respectively. In particular, the standard deviation $\sigma$ may be noticeably influenced by salient
trajectories. Indeed, it tends to make the $q_i$ smaller and it turns out that only fewer $q_i$ are finally above the threshold $\lambda$ given by the \textit{p}-value. As a consequence, very few false detections occur but at the cost of misdetections of salient
trajectories. We would have to decrease the $p$-value. This is
confirmed by experiments reported in Appendix~\ref{sec:alt_lambda}.
Then the \textit{p}-value should be adapted to the considered
scenario, which amounts to directly set the threshold $\lambda$. The
\textit{p}-value scheme is not effective in our case.

A way to overcome this problem would be to take another definition of the $q_i$ descriptors that would not depend on the presence of
salient trajectories. A first idea would be to compute $\bar{d}$ and $\sigma$ once and for all on normal trajectories of the validation set for each scenario. This way, the dependence on the presence of salient trajectories at test time vanishes. Another idea would be to make a robust estimation of $\bar{d}$ and $\sigma$ on the scenario at test time. However, experiments conducted on the datasets described in Section~\ref{sec:datasets}, showed that results obtained for both alternatives were not as good as the ones obtained with the initial $q_i$.

\subsubsection{Hard thresholding}

In practice, we have adopted the following approach. To set $\lambda$ for a given version of the trained network and a given dataset, we probed a range of values for $\lambda$. More specifically, we test candidate values over the interval [0,5] sampled every 0.05. Then, we select the one providing the best F-measure over the validation set associated to the given experiment. The choice $\lambda = 5$ as upper bound of the interval corresponds to 5 times the standard deviation with the definition of the $q_i$ descriptors. It is enough to embrace very salient elements.

\section{{ Datasets, Training and Saliency}}
\label{sec:datasets}

In this section, we present the datasets used to train and evaluate
our method. We have built a synthetic dataset of trajectories called
STMS (Synthetic Trajectories for Motion Saliency estimation). We have
also used the dataset made available by \cite{Alahi2014}, comprising
pedestrian trajectories acquired in a railway station with a set of
cameras over a long time period. We denote it as RST dataset (RST
standing for Railway Station Trajectories). { However, we had to organize it properly to be able to apply trajectory-saliency detection on it.}

\subsection{STMS dataset}
\label{sec:STMS}

Our STMS dataset of synthetic trajectories for motion saliency, contains three trajectory classes: straight lines, trajectories with sharp turns and circular trajectories. A noise of small magnitude is added to the successive trajectory positions to increase the variability of the trajectories.

The construction of synthetic datasets is a common practice in many computer vision tasks involving deep neural networks, for instance for optical flow estimation. Its interest is two-fold. It allows a large-scale objective experimental evaluation, since by nature, a synthetic dataset is not limited and ground truth is immediately available. It enables to pre-train the network with a large set of examples, making it more efficient on real data.

Each scenario is composed of trajectories of the same kind (e.g., straight line) and with similar parameters (velocity, initial motion direction, etc.). The number of positions is comprised between 20 and 60 for each trajectory. The velocities vary within $[5, 20]$ depending on the scenarios. The initial direction is randomly taken in $[0, 2\pi]$. For circular trajectories, the angular velocity is constant and lies in $[-0.10, 0.10]$ radians per time step. For trajectories with sharp turns, the rotation angle is chosen in
$[-\frac{\pi}{2}, -\frac{\pi}{6}]~\cup~[\frac{\pi}{6},
\frac{\pi}{2}]$. The initial position of the trajectories is set to the origin (0,0). With the addition of the turn time instant for trajectories with sharp turns, we have defined the set of parameters to specify each trajectory class. For a given scenario, normal trajectories should be similar. As a consequence, only small variations are allowed for the trajectory parameters in a given scenario. Yet, to avoid too uniform trajectories, a random noise is added to each trajectory. The random addition $(x_n(t), y_n(t))$ to the position coordinates at time $t$ depends on $(x_n(t-1), y_n(t-1))$ to ensure that the trajectory remains smooth.

\begin{figure}[!htbp]
  \centering
  \begin{tabular}[!htbp]{|c|c|c|}
    \hline
    \includegraphics[width=0.28\linewidth]{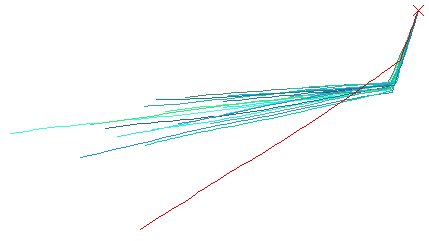}
    & \includegraphics[width=0.28\linewidth]{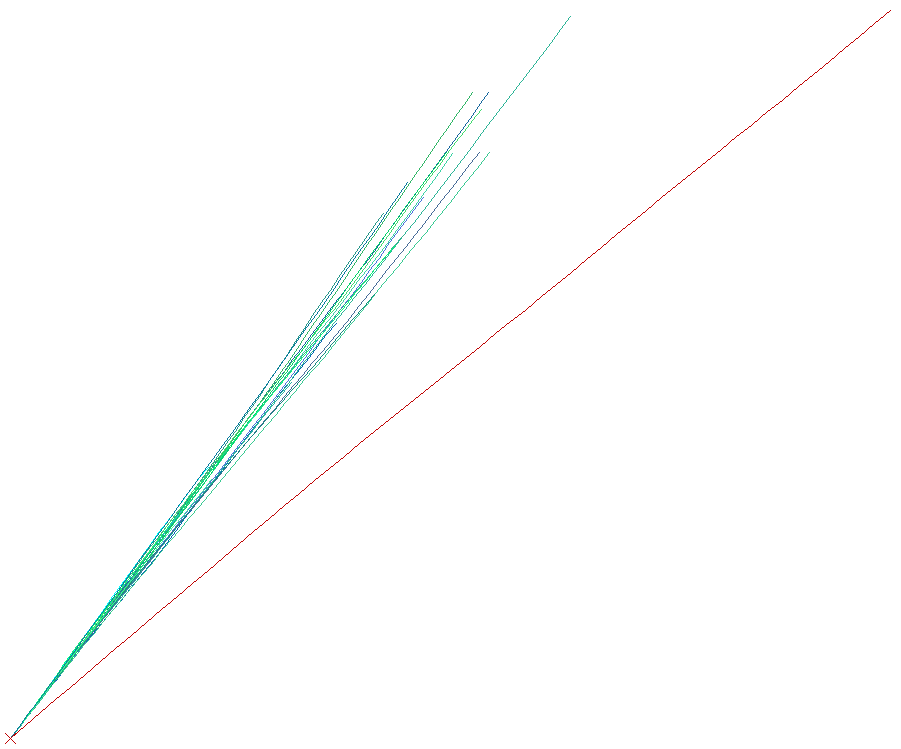}
    & \raisebox{-2pt}[0.132\textwidth]{\includegraphics[width=0.28\linewidth]{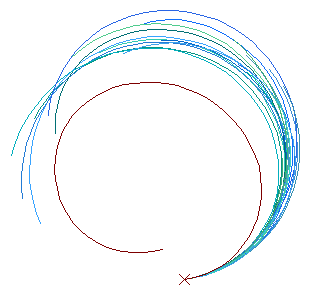}} \\ \hline
    a) Sharp turn & b) Straight line & c) Circular \\ \hline
  \end{tabular}
  \caption{Examples of synthetic trajectories for the three classes,
    starting from the same origin. Salient trajectories are drawn in
    red, normal trajectories in green to blue.}
  \label{fig:illustration_synthetic}
\end{figure}

To come up with a difficult enough task, salient trajectories are of
the same class as normal trajectories. The difference will lie in the
parametrisation of the salient trajectories. The parameters are chosen
so that the salient trajectory should be not too different from normal
trajectories, but still distinct enough to prevent any visual
ambiguity about its salient nature. An illustration is given in
Figure~\ref{fig:illustration_synthetic}.

The trajectories for training are generated on the fly. We generate
scenarios composed of 10 normal trajectories. An additional salient
trajectory may be added with a probability of 0.5. Accordingly, we
build batches made of 6 scenarios each, for a total of 60 to 66
trajectories.

The validation set and the test set are composed of 500 scenarios
each. For these two sets, each scenario is composed of 20 normal
trajectories. An additional salient trajectory is included with a
probability of 0.5. We end up with a mean rate of salient trajectories
of 2.5\%. These two different configurations for the training and the
test were chosen to challenge our method. A higher ratio of saliency
is expected to make the training more difficult. On the other hand,
rare salient trajectories are more difficult to find by chance during
the test.

\subsection{RST dataset}
\label{subsec:RST_dsc_tra}

We now describe the RST dataset of real trajectories. We start with a
general description of the dataset, before presenting the training
procedure we applied to the network for this dataset.

\subsubsection{Description of the RST dataset}

\begin{figure}[!htbp]
  \centering
  \begin{tabular}{cc}
    \includegraphics[width=0.465\linewidth]{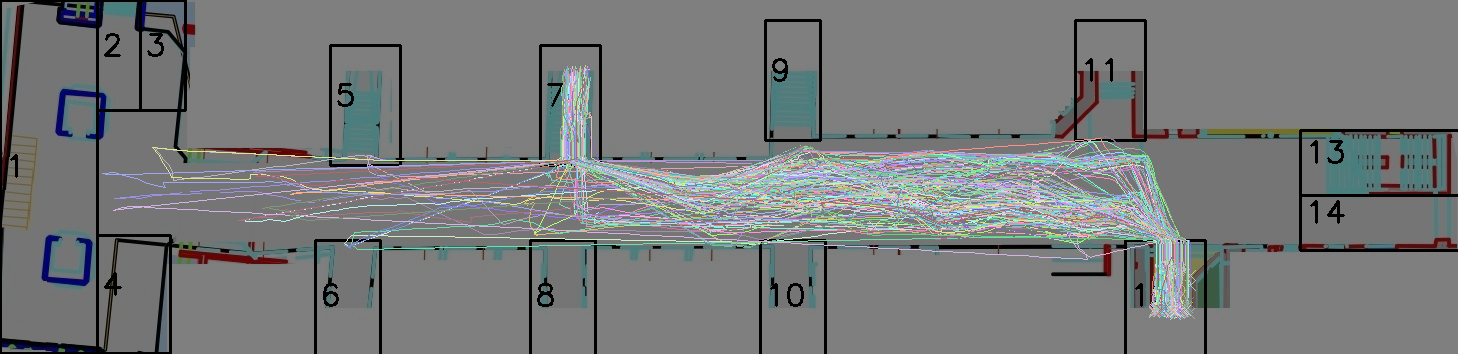}
    & \includegraphics[width=0.465\linewidth]{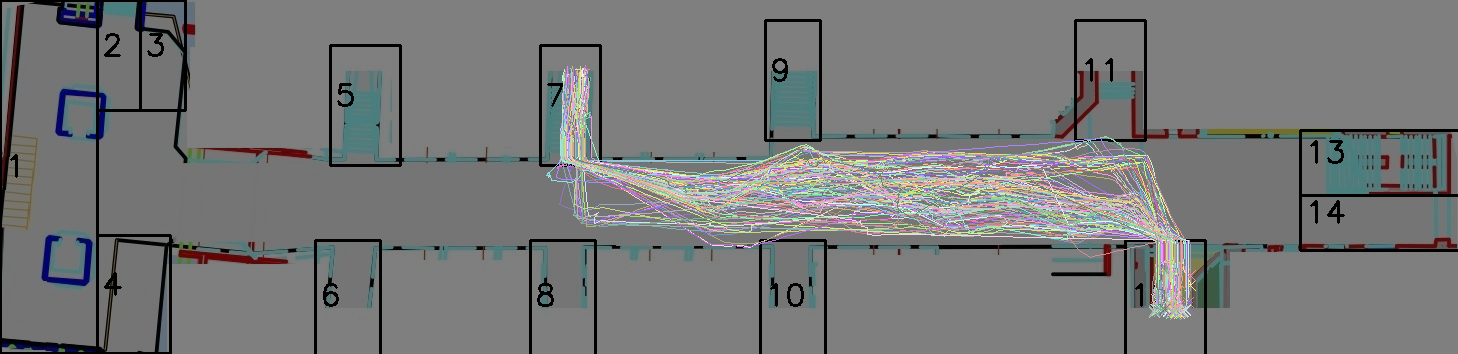} \\
    \includegraphics[width=0.465\linewidth]{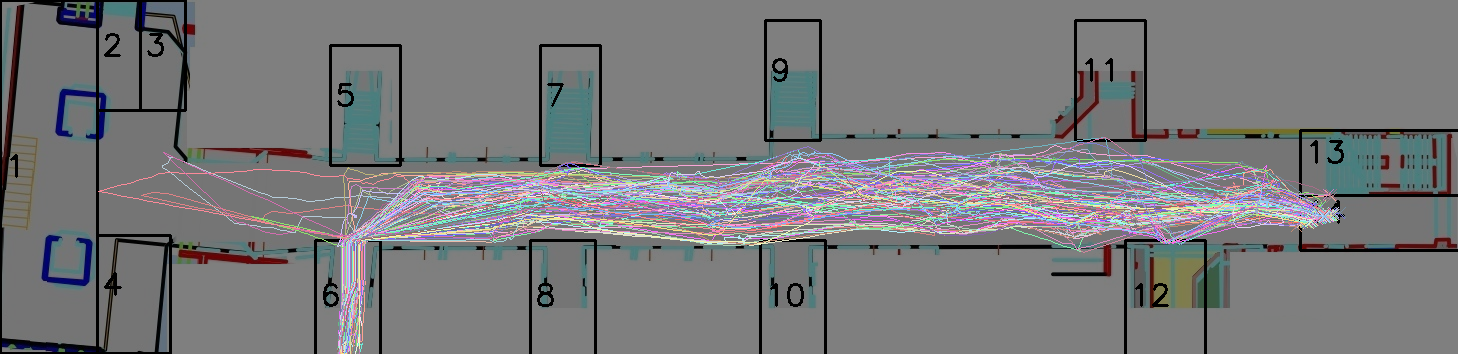}
    & \includegraphics[width=0.465\linewidth]{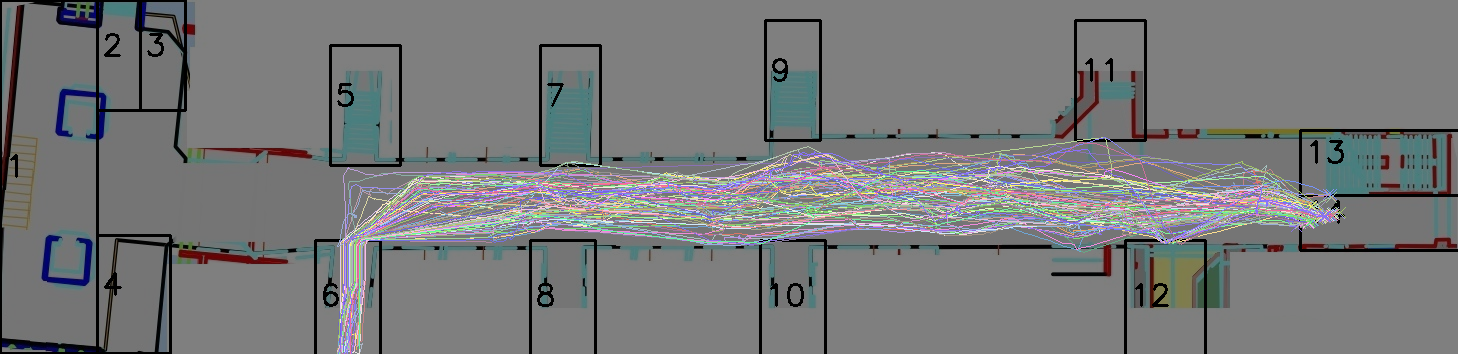} \\
    \includegraphics[width=0.465\linewidth]{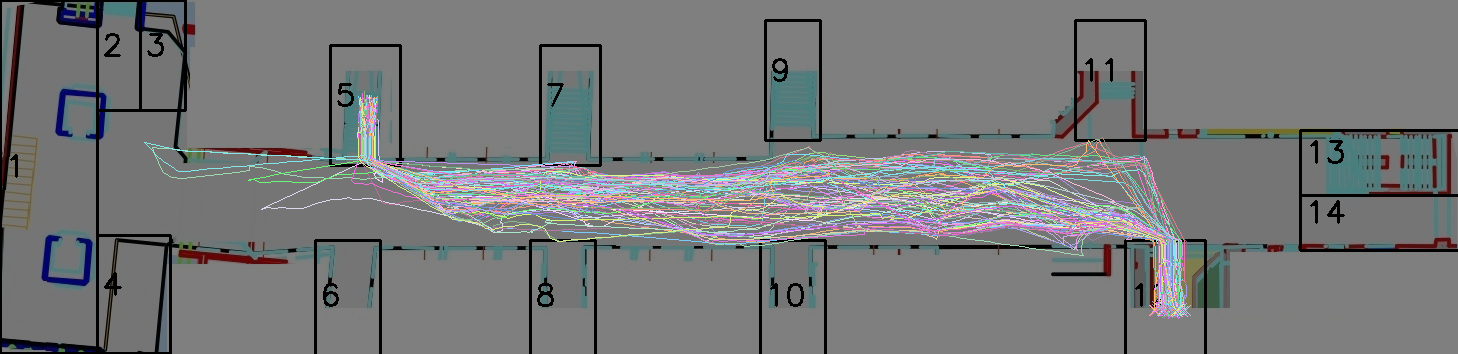}
    & \includegraphics[width=0.465\linewidth]{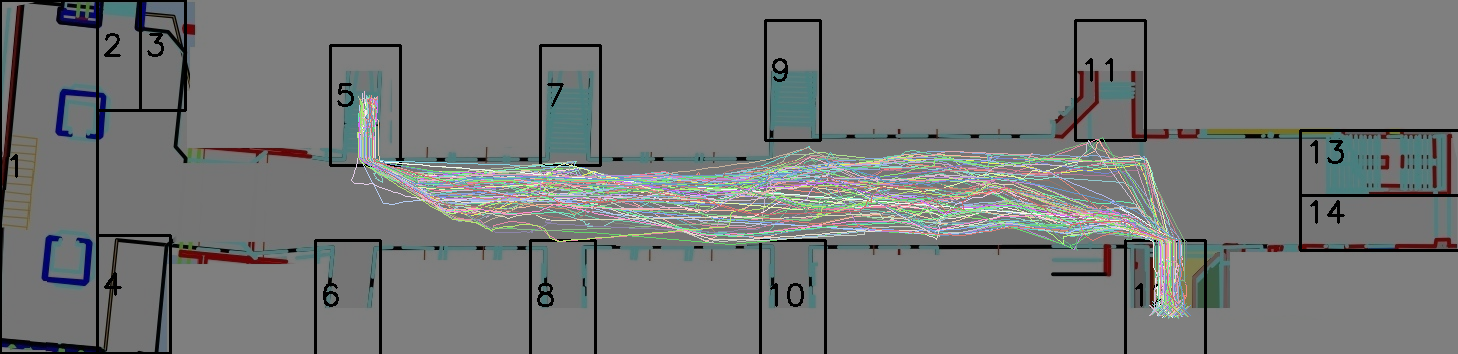} \\
  \end{tabular}
  \caption{Illustration of the RST dataset (corridor PIW, numbers
    correspond to gates) and of the pre-processing step used to get
    homogeneous motion patterns. { Display of trajectories
    before removing the "erratic" ones on the left, after removing them on the
    right}.}  \label{fig:pre_proc_batch}
\end{figure}

The second dataset is constituted of publicly available real
pedestrian trajectories, computed with a tracking system installed in a train station \cite{Alahi2014}. The tracking system takes videos recorded by a network of cameras. The videos were processed to extract trajectories, which are projected on the 2D ground plane of the train station. More specifically, we use the trajectories extracted from one underground corridor, denoted in the following as PIW, with gates on its sides and its extremities (the corridor is displayed in Figure~\ref{fig:pre_proc_batch} with trajectory samples). The acquisition was performed during 13 days in February
2013, and resulted in a total of 115,245 trajectories. Training data
are sampled from the 95,458 trajectories acquired during the 11 first days. We use trajectories sampled from the 19,787 trajectories
acquired during the last two days for validation and test.

Compared to the STMS dataset, the trajectories of the RST dataset are
longer, and the mean displacement between two points is larger.  Very
long trajectories are challenging for recurrent networks such as the
one we employ. To overcome it, we subsampled RST trajectories by
keeping one every five points (consequently multiplying the observed
displacement magnitude by a factor five). We also divided the
coordinates by 100 to get displacement magnitudes closer to the
synthetic case. The full trajectories are recovered by
re-identification from one camera to the next. This means that a given
trajectory may be interrupted for some time, before being continued at
a different point. To avoid these irregularities, we pre-processed the
trajectories by splitting them when unexpectedly large displacements
are observed.

For a better stability of the training, we found helpful to make all
the trajectories start at the same origin $(x_0, y_0)$. It can be
viewed as a translation-invariant requirement with respect to the
location of the trajectory in the 2D-space.

\subsubsection{Training procedure for the RST dataset}
\label{RTS-training}

We pre-trained our network with the synthetic trajectories.
{ For the training procedure on the RST dataset, we have to organize the data to apply the consistency constraint, since the RST dataset involves many different motion patterns. Accordingly, we build scenarios composed of
trajectories sharing the same entry and exit; this information is directly available in the trajectories files. It may happen that between a given entry and exit, people follow a more random (or even "erratic") path. They usually go straight from the entry to the exit, but some individuals may make a detour to another location, as illustrated in
Figure~\ref{fig:pre_proc_batch}. We prefer to remove these "erratic" paths from the training scenarios.} To this end, we computed the median length of
trajectories associated to a given entry/exit pair. Only
trajectories with a length deviating from less than 10\% of the median
length were kept. We first built elementary scenarios of 8
trajectories by following the above-mentioned procedure. To get
training batches of about 60 trajectories as in the synthetic dataset,
we grouped 8 elementary scenarios from several entry/exit pairs to
construct a batch. It improves the diversity of data at each training
iteration.

We consider three levels of consistency, respectively of $\beta=10^5$, strong consistency, $\beta=10^3$, low consistency, and $\beta=0$, no consistency. The three network variants trained as explained above with these three consistency levels will be denoted respectively $V\beta_5$, $V\beta_3$ and $V\beta_0$ in the sequel. For this training procedure, 43,079 trajectories from the corridor PIW are included in the training set.
In addition, we removed some trajectories, such as the ones
starting in the middle of the corridor, or the ones following rare
paths. The training of our method variants is done once for all on this training set and will be used for all the experiments reported in Section \ref{subsec:res_gare} apart from a specific one as explained later.

\subsection{{ RST dataset organization for trajectory-saliency detection}}
\label{sec:eval_setting}

The goal is now to evaluate our method on real trajectory data. The RST dataset is an appealing dataset since it contains a large set of trajectories of different patterns. However, it was built for crowd analysis, not specifically for trajectory-saliency detection. There is no predefined saliency ground truth. Therefore, we have to design our own saliency experiments from the available data. On one hand, we have defined two evaluation dataset from the RST dataset:

\begin{itemize}
\item RSTE-LS is an organized version of the RST test set based on the entry-exit gate pairs, in order to evaluate our method on a somewhat large scale,
\item RSTE-CP is a set of limited size needed to carry out comparative evaluation with other existing methods requiring specific training.
\end{itemize}

On the other hand, we have defined three kinds of trajectory saliency:
\begin{itemize}
\item Trajectories with different entry and/or exit gates (called DT-saliency in the sequel),
\item Trajectories that do not go directly from entry to exit, that is, the "erratic" ones (called ET-saliency),
\item Faster trajectories (called FT-saliency).
\end{itemize}

\subsubsection{Evaluation set RSTE-LS}
\label{rste-ls}

{ The trajectories of the dataset are organized by entry-exit gate pairs. To ensure that the normal trajectories attached to a given entry-exit pair, follow a similar motion pattern, we remove the "erratic" ones as done for the training procedure. For DT-saliency, trajectories salient with respect to normal trajectories of a given entry-exit pair, have either a different entry or a different exit, or both. Obviously, the same trajectory may play two different roles, either the normal one, or the salient one, depending on the entry-exit pair considered.}

We built a RSTE-LS test set comprising 2275 normal trajectories. They are
divided into 45 scenarios, corresponding to 45 different entry/exit
pairs.
{ First, we have studied the DT-saliency on RTSE-LS. In each scenario, salient trajectories are trajectories corresponding to a different entry/exit pair.} We have considered three degrees of saliency. The highest the saliency degree, the easiest the trajectory saliency detection.

\begin{itemize}
\item In the easiest case, salient trajectories are randomly drawn
  from entry/exit pairs different than the one of the normal trajectories. Furthermore, only entry/exit pairs leading to distinct motion pattern are allowed. For instance, if normal trajectories go from gates 7 to 8, salient trajectories cannot go   from gates 9 to 10 since the trajectories would be parallel, and consequently, of the same motion pattern once translated to the common origin (see Figure~\ref{fig:pre_proc_batch}).
\item In the medium case, salient trajectories share either the
  entrance or the exit with normal trajectories. The other gate of
  salient trajectories is chosen two positions after or before the
  other gate of normal trajectories on the same side of the
  corridor. For instance, if a normal trajectory goes from gates 12 to
  9, a salient trajectory may go from gates 12 to 5, gate 5 being two
  positions after gate 9.
\item The most difficult case is built similarly as the medium
  case. The difference is that the gate of a salient trajectory that
  differs from the gate of normal trajectories is only one position
  apart, that is, the next gate on the same side of the corridor. For
  instance, for a normal trajectory going from gates 12 to 9, a
  salient trajectory may go from gates 12 to 7.
\end{itemize}

In the following, these cases will be designated with their degree of
saliency respectively qualified as high, middle and low.
In addition to the saliency degree, we will consider different ratios
of salient versus normal trajectories. Ratios of 5\%, 10\% and 15\%
will be tested.

{ Second, we have studied the FT-saliency on RSTE-LS.
We set up this experiment by sub-sampling a small group of trajectories for each entry-exit gate pair, which is a way to implement faster paths. They act as the salient trajectories in each entry-exit gate pair.}~\\

\subsubsection{Evaluation set RSTE-CP}
\label{subsec:res-b}

\begin{figure}[!htbp]
  \centering
  \begin{tabular}{ccc}
    \includegraphics[width=0.8\linewidth]{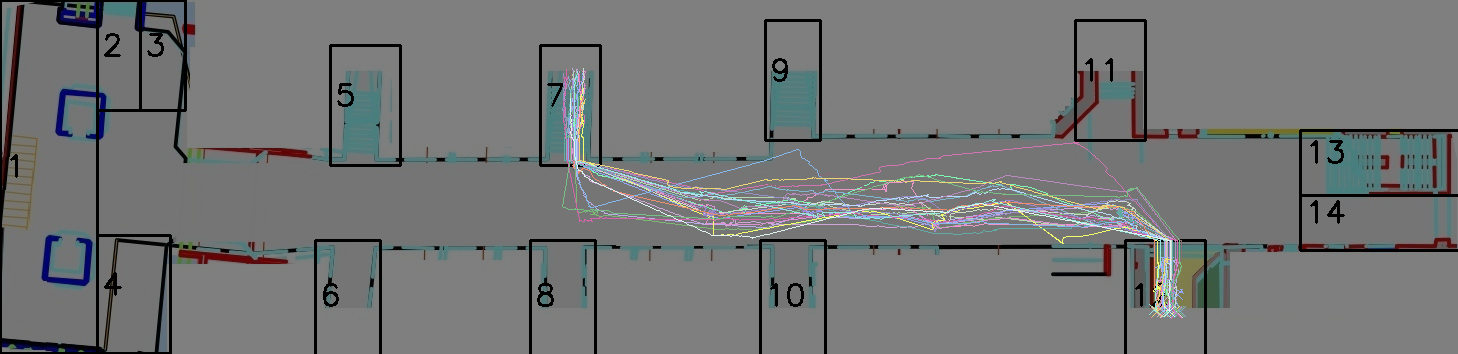}   \\
    \includegraphics[width=0.8\linewidth]{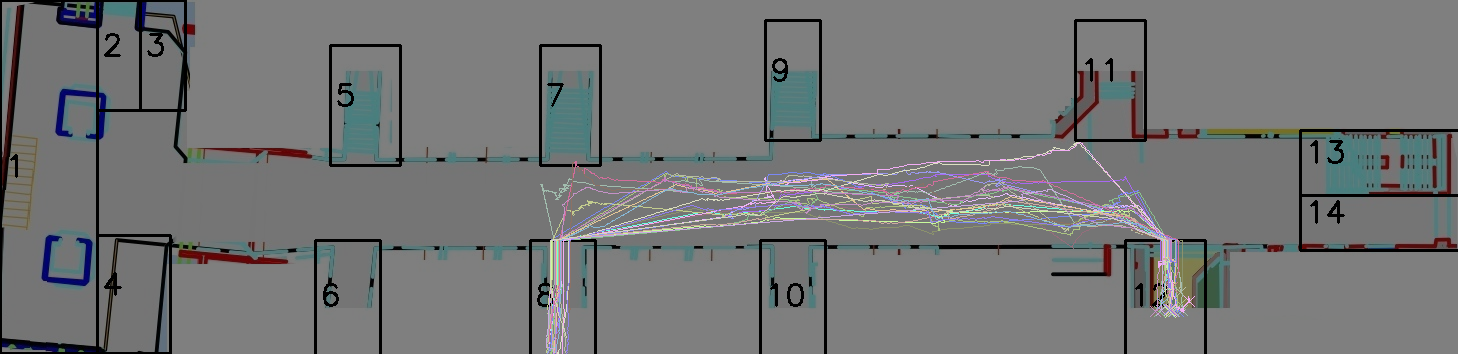}   \\
    \includegraphics[width=0.8\linewidth]{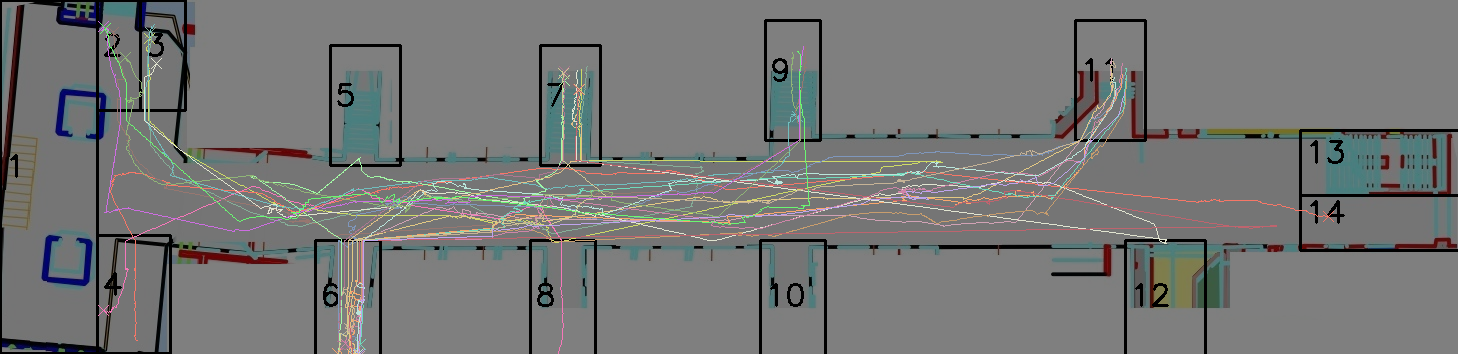} \\
  \end{tabular}

  \caption{{ From top to bottom: 25 normal trajectories from the G12-7 subset, 25 normal trajectories from the G12-8 subset (the two subsets forming the evaluation set RSTE-CP), 25 salient trajectories with different entry/exit pairs used for the DT-saliency experiment.}}  \label{fig:traj_gare}
\end{figure}

{ We built a second evaluation set RSTE-CP to be able to compare with existing methods. Indeed, some of them need to be trained on normal
data exclusively. As a consequence, for these methods, each scenario requires its own training set, and training is performed for every scenario. To ensure it and for a fair comparison, we double-checked manually the sets of normal trajectories for training, which necessarily led to an evaluation set of limited size. The normal trajectories are
checked by hand to make sure they all follow a consistent motion
pattern. Therefore, RSTE-CP involves only two entry/exit pairs, instead of 45 for RSTE-LS.

We compare the three variants of our method to four existing methods, DAE \cite{Roy2018}, ALREC \cite{Roy2019}, TCDRL \cite{Yao2017} and AET \cite{Ma2018}. The comparison also comprises a baseline taking into account only the trajectory length. The baseline classifies a trajectory as salient if the trajectory length deviation is more than a given percentage of the median trajectory length. In fact, this is nothing more than the way we proceed to remove "erratic" trajectories in each entry-exit gate pair for training our method on the RSTE-LS set. We applied the baseline with two length rates: 10\% and 15\%.}

More precisely, 200 normal trajectories are selected coming from gate 12 and going to gate 7. We name them as the G12-7 subset. 200 other trajectories are
selected coming from gate 12 and going to gate 8. They are designated
as the G12-8 subset. Samples are depicted in
Figure~\ref{fig:traj_gare}.  For each subset, half of the trajectories
is used for training and the other half is used for test. The training
subset is denoted as TrG. The training set TrG is used for DAE \cite{Roy2018}, ALREC \cite{Roy2019}, and AET \cite{Ma2018} in the comparative experiments.~\\

{ First, we designed a trajectory-saliency experiment on RSTE-CP based on the DT-saliency, that is, salient trajectories are associated to different entry/exit pairs. 100 trajectories are drawn from other entry-exit pairs than G12-7 and G12-8, according to the easiest case defined above in subsection \ref{rste-ls}. They will serve as salient trajectories w.r.t. to both G12-7 and G12-8 subsets.} For each G12-7 and G12-8 test subsets, we have built 20 scenarios, each with 100 normal and 5 salient trajectories, leading to a saliency ratio of 5\%.

\begin{figure}[!htbp]
  \centering
  \begin{tabular}{ccc}
    \includegraphics[width=0.8\linewidth]{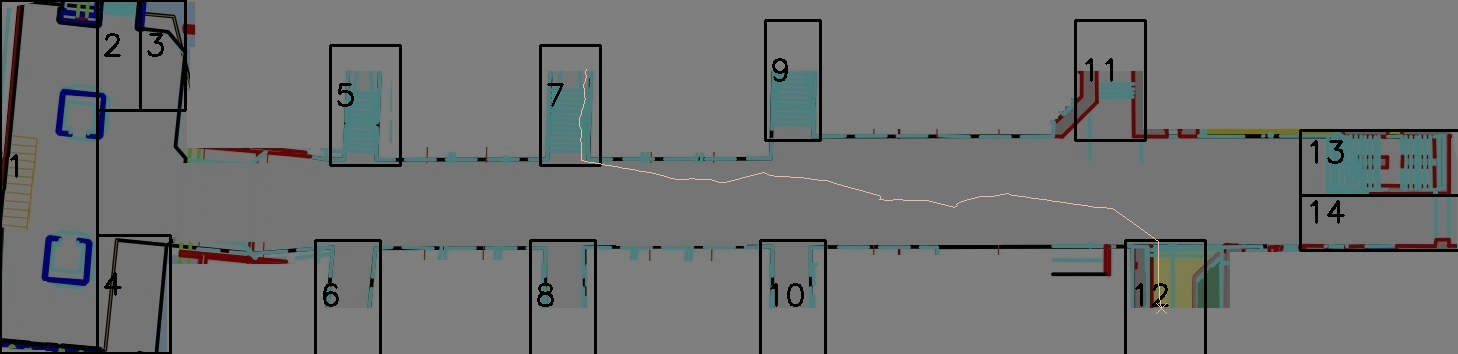}   \\
    \includegraphics[width=0.8\linewidth]{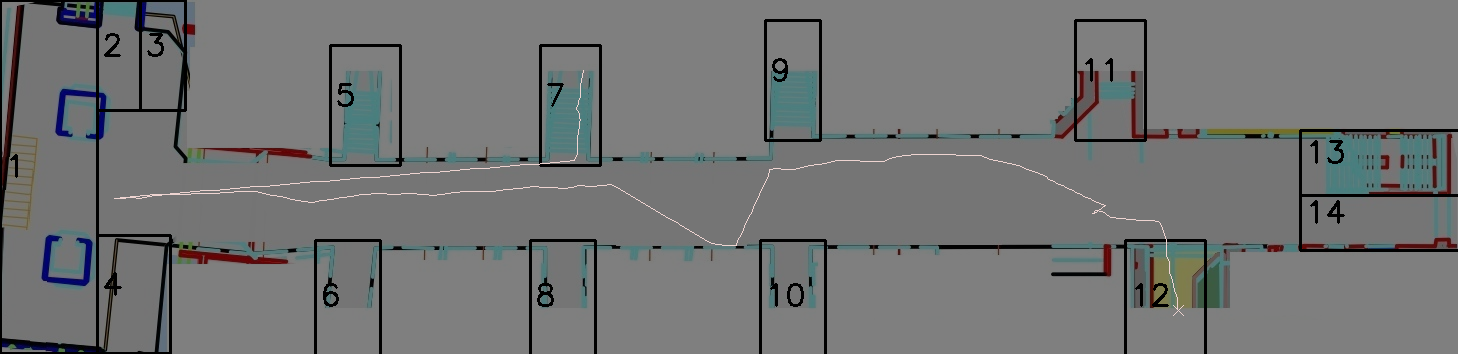}   \\
  \end{tabular}
  \caption{{ A normal trajectory (top) and a salient trajectory (bottom) for the ET-saliency experiment applied to the RSTE-CP set.}}
  \label{fig:traj_gare_TII}
\end{figure}

{ Second, we have designed another trajectory-saliency experiment on RSTE-CP based this time on ET-saliency. In each subset, G12-7 and G12-8, the salient trajectories are now those that do not go straight from the entry to the exit, the "erratic ones" (see Figure~\ref{fig:traj_gare_TII} for an illustration). Actually, it is a special saliency situation, because the trajectory length is sufficient to characterize ET-saliency. To be sure that all trajectories are properly assigned to the normal or salient class in the ground truth, the labelling is performed manually for this experiment. In contrast to the RTS training procedure described in subsection \ref{RTS-training}, we include erratic trajectories in the training of our method in this specific experiment for a fair comparison of the three variants. For each gate pair, the test set will include a total of 155 trajectories divided into 137 normal ones and 18 salient ones.}

\section{Experimental results}
\label{sec:res}

In this section, we report the experiments we carried out on the
synthetic STMS dataset and the real RST dataset.

\subsection{Implementation details}
\label{sec:dtr}

The network was implemented with the PyTorch framework
\cite{Paszke2019}. For training, we used the Adam algorithm
\cite{Kingma2014}. We set the learning rate by training the
auto-encoder network on the STMS dataset. A learning rate of $10^{-4}$
provided a good compromise between speed and convergence and was then
retained.

Parameter $\beta$ of the loss function~(\ref{eq:unsup_total_loss})
defines the balance between the reconstruction and consistency
constraints. It was set as follows. On one hand, a too low value of
$\beta$ would make the consistency constraint negligible. On the other
hand, a too high value of $\beta$ would prevent the auto-encoder from
correctly reconstructing trajectories. In practice, we set
$\beta=10^5$ for the experiments on the STMS dataset. For the RST
dataset, we considered values of $\beta=10^5$, $\beta=10^3$ and
$\beta=0$, as mentioned in Section~\ref{subsec:RST_dsc_tra}.

Regarding the computation time, the forward pass takes 0.1 second and
the backward pass 0.5 second, for a batch of 10 trajectories
comprising 40 positions, on a machine with 4 CPU cores of 2.3 GHz. We
let each network variant train three weeks.

\subsection{Results on the STMS synthetic dataset}
\label{subsec:syn}

We first evaluate our trajectory saliency method on the STMS synthetic
dataset presented in Section~\ref{sec:STMS}. Our first goal is to
assess the quality of the trajectory reconstruction by the
auto-encoder. The quality score $r$ will be defined as the average
reconstruction error $\bar{e}$, divided by the average displacement
$\bar{v}_n$ over the whole dataset:

\begin{equation}
  \label{eq:r_rec_metric}
   r = \frac{\bar{e}}{\bar{v}_n}, \mbox{~~~with}
 \end{equation}
\begin{equation}
 \label{eq:rec_error}
  \bar{e} = \frac{1}{N} \sum_{\mathcal{T}_i} \frac{1}{t_{final} - t_{init} + 1}  \sum_{t=t_{init}}^{t_{final}} \sqrt{(x_t^i - \hat{x}_t^i)^2 + (y_t^i - \hat{y}_t^i)^2}.
  \end{equation}
N is the number of trajectories in the dataset. $r$ is defined this way to be dimensionless.

\begin{figure}[!htbp]
  \centering
  \begin{tabular}[!htbp]{|c|c|c|}
    \hline
    \raisebox{-2pt}[0.115\textwidth]{\includegraphics[scale=0.076]{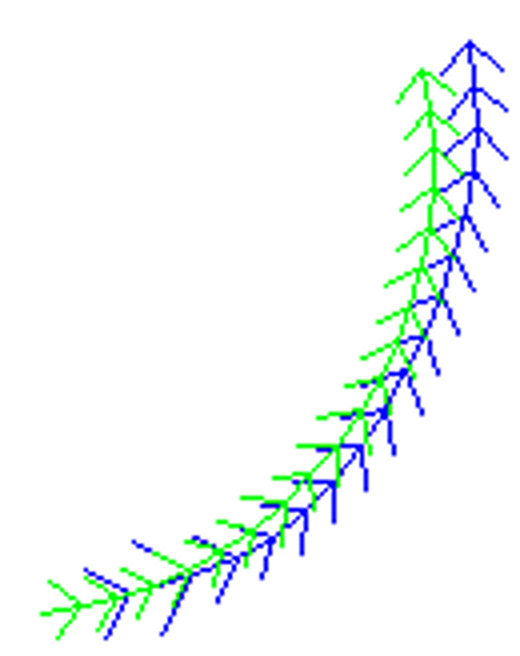}}
    & \raisebox{-2pt}[0.115\textwidth]{\includegraphics[scale=0.071]{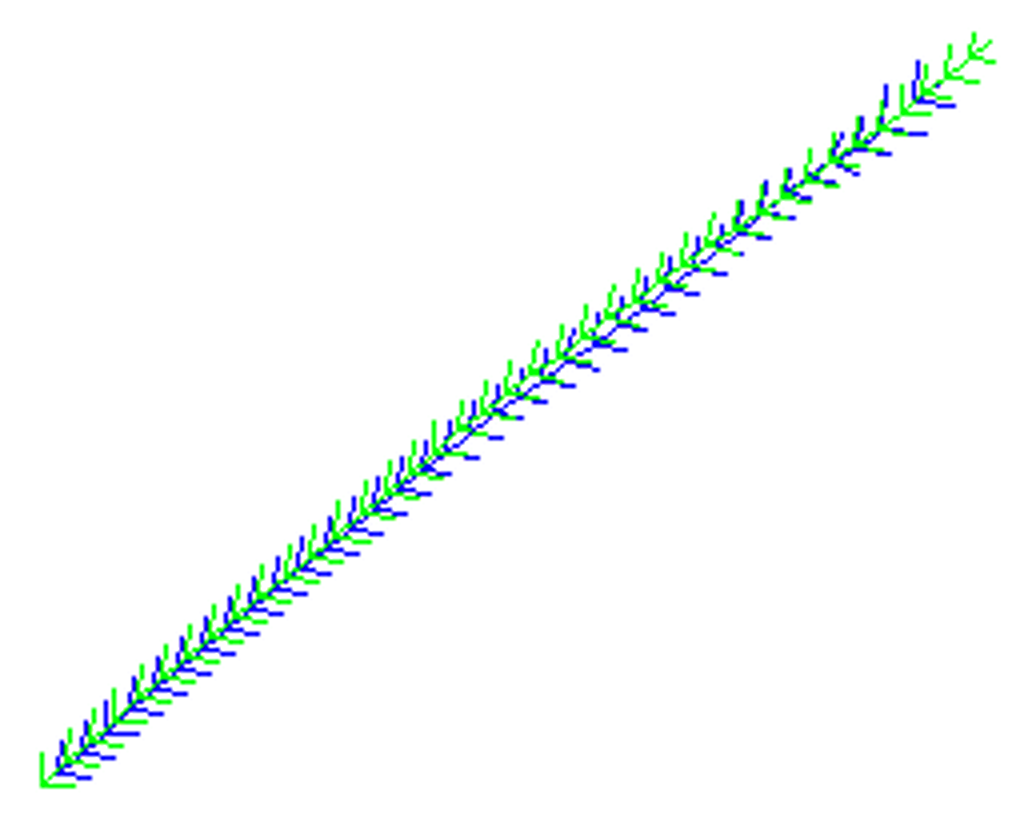}}
    & \raisebox{-2pt}[0.115\textwidth]{\includegraphics[scale=0.071]{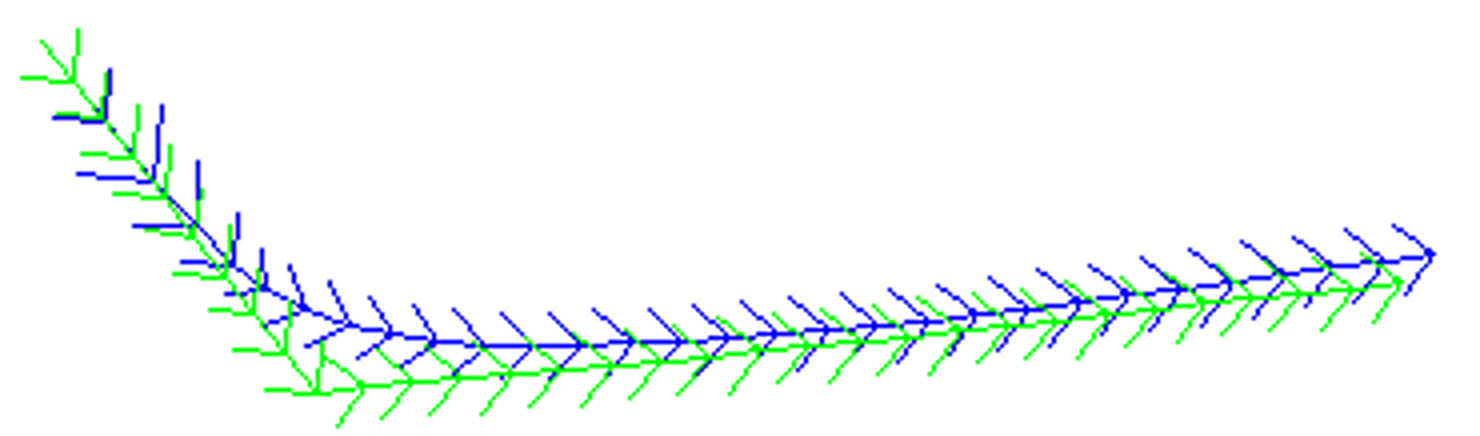}} \\ \hline
  \end{tabular}
  \caption{Three examples of synthetic trajectories (in green) and their reconstructed counterpart superimposed (in blue) for the three classes of trajectories. They nicely match, demonstrating an accurate reconstruction.}
  \label{tab:rec_syn}
\end{figure}

Our network is trained on the STMS dataset with the procedure
described in Section~\ref{sec:dtr}. For this dataset, we
consider only one variant with $\beta$ set to $10^5$. We got a score
$r=0.62$ for this dataset. We plot samples of reconstructed
trajectories in Figure~\ref{tab:rec_syn}.

The second goal is to assess the trajectory saliency detection
performance. We compute precision, recall and F-measure on the class of salient trajectories only, because they are the ones we are interested in. The rationale of this choice is to avoid a strong bias on performance due to overwhelming normal trajectories. The F-measure is the harmonic mean of precision and recall:
\begin{equation}
  \label{eq:f_measure}
  \mbox{F-measure} = 2 \cdot \frac{\mbox{precision} \cdot \mbox{recall}}{\mbox{precision} + \mbox{recall}}
\end{equation}

Let us recall that the STMS dataset was constructed with 2.5\% of salient trajectories. As shown in Table~\ref{tab:res_STMS}, we reached a F-measure equal to 0.89 for this dataset. It proves that we were able to correctly find even subtle trajectory saliency.

To highlight the contribution of the consistency constraint, we
conducted an ablation study of our method. More specifically, the
consistency constraint is removed, letting the auto-encoder
reconstruction constraint drive the latent code estimation. The
reconstruction score is $r=0.58$, and as expected it is slightly
better than the one $r=0.62$ obtained with the additional consistency
constraint. However, the F-measure collapses to 0.26 as shown in
Table~\ref{tab:res_STMS}. This clearly demonstrates the importance of
the consistency constraint. Our intuition is that without the
consistency constraint, the network employs all the available degrees
of freedom to encode the trajectory pattern including any small
variation. With the consistency constraint, the network is more
focused on correctly representing the overall motion pattern of each
trajectory, which allows us to better distinguish normal and salient
trajectories.

\begin{table}[!htbp]
  \centering
  \footnotesize
  \begin{tabular}{|c||c|c|c|}
    \hline
                        & Precision & Recall & F-measure \\ \hline
    Our full method     & 0.91      & 0.87   & 0.89      \\ \hline
    Without consistency & 0.22      & 0.31   & 0.26      \\ \hline
  \end{tabular}
  \caption{Performance of our method evaluated on the STMS dataset,
    with and without the consistency constraint. We computed
    precision, recall and F-measure w.r.t. the salient trajectory
    class.}
  \label{tab:res_STMS}
\end{table}

\subsection{ { Results on the evaluation sets of the RST dataset}}
\label{subsec:res_gare}

{ In this section, we present results obtained on the two sets of trajectories RSTE-CP and RSTE-LS issued from the RST dataset of real pedestrian trajectories acquired in the train station.}~\\

\subsubsection{{ Comparative results on the RSTE-CP evaluation set with DT-saliency}}

{ We report the comparative evaluation carried out on the RSTE-CP evaluation set for two kinds of trajectory saliency, DT-saliency and ET-saliency.
This experimental comparison involves our three method variants and four existing methods, namely, DAE \cite{Roy2018}, ALREC \cite{Roy2019}, TCDRL \cite{Yao2017} and AET \cite{Ma2018}. In addition, we include the baseline method introduced in subsection \ref{subsec:res-b}.}

\begin{table}[!htbp]
  \centering
  \begin{tabular}{|c||c|c|c||c|c|c|}
    \hline
    RST subset                 & \multicolumn{1}{c}{}  & \multicolumn{1}{c}{G12-7}  & \multicolumn{1}{c||}{} & \multicolumn{1}{c}{}  & \multicolumn{1}{c}{G12-8}  & \multicolumn{1}{c|}{} \\ \hline

    Method                    & P    & R    & FM   & P    & R    & FM   \\ \hline
    DAE \cite{Roy2018}        & 0.83 & 0.32 & 0.46 & 0.81 & 0.32 & 0.46 \\ \hline
    ALREC \cite{Roy2019}      & 0.72 & 0.43 & 0.54 & 0.71 & 0.45 & 0.55 \\ \hline
    TCDRL-       & & & & & & \\
    KMeans \cite{Yao2017}      & 0.05 & 0.54 & 0.10 & 0.08 & 0.53 & 0.14 \\
    \hline
    TCDRL-C \cite{Yao2017}      & 0.72 & 0.24 & 0.36 & 0.54 & 0.52 & 0.53 \\
    \hline
    AET \cite{Ma2018}      & \textbf{1.0} & \textbf{1.0} & \textbf{1.0} & \textbf{1.0} & \textbf{1.0} & \textbf{1.0} \\
    \hline
    Baseline 10\%      & 0.32 & 0.94 & 0.48 & 0.47 & 0.93 & 0.62 \\
    \hline
    Baseline 15\%      & 0.82 & 0.89 & 0.85  & 0.69 & 0.88 & 0.77 \\
    \hline
    Ours $V\beta_5$           & \textbf{1.0}  & 0.98 & 0.99 & \textbf{1.0}  & 0.89 & 0.94 \\ \hline
    Ours $V\beta_3$           & \textbf{1.0}  & 0.98 & 0.99 & \textbf{1.0}  & 0.98 & 0.99 \\ \hline
    Ours $V\beta_0$           & \textbf{1.0}  & 0.99 & 0.99 & \textbf{1.0}  & 0.99 & 0.99 \\ \hline

  \end{tabular}
  \caption{Comparative results for the RSTE-CP evaluation set with DT-saliency, on the G12-8 and G12-7 test subsets. P, R and FM denote respectively precision, recall and F-measure computed w.r.t. the salient trajectory class. Best results are in bold.}
  \label{tab:res_gare}
\end{table}

First, we deal with the DT-saliency experiment.
For a fair comparison, the decision threshold involved in each method is set as best as possible. Our method variants are trained as described in
Section~\ref{subsec:RST_dsc_tra}. The decision threshold $\lambda$ is
set over a validation set with low saliency degree and saliency ratio
of 5\%.

DAE and ALREC methods estimate saliency by training an auto-encoder on
normal data only, and by considering that trajectories poorly
reconstructed are salient. Then, for a fair comparison, we trained
each of the two networks on the training subset of G12-7 and on the
training subset of G12-8 respectively according to the experiment
conducted. The decision threshold is set as proposed by the authors
for DAE and ALREC. We then estimated trajectory saliency first
on the G12-7 subset and then on the G12-8 test subset. The training
subsets include 100 trajectories per scenario, which is more than the
16 to 31 trajectories per scenario that the authors considered in
their own work. Due to the higher number of trajectories, we did not
include data augmentation.

The TCDRL method represents trajectories with a constant-length code. This representation is obtained directly by training a recurrent network
on the test trajectories. TCDRL was originally designed to produce input to clustering algorithms. To apply TCDRL to the trajectory saliency detection task, we adapted it and we defined two alternatives. The first option consists in using the \textit{k}-means clustering algorithm with two classes. We then state that the class with fewer elements is the salient class.  However, there is a risk that the large imbalance between the normal and salient classes makes this first option unstable. It motivates the following alternative. The second option exploits our decision test, but we applied it to the code produced by TCDRL. We took 101 values of $\lambda$ regularly sampled in $[0,5]$, and we selected the one that provided the best performance for TCDRL on the test set.

{  For the AET method, we used the same training sets than for DAE and ALREC. It means that 100 RNNs are trained. AET estimates an anomaly score, that needs to be converted into a binary prediction. As for TCDRL, we sampled 101 thresholds and selected the one that provided the best performance on the test set.}

Results for DT-saliency on the two subsets of the RSTE-CP evaluation set are given in Table~\ref{tab:res_gare}. { We observe that our method variants and the AET method provide almost or even perfect results for this experiment. They outperform the other methods.}~\\

\subsubsection{{ Comparative results on the RSTE-CP evaluation set with ET-saliency}}

{ We now report comparative results with the ET-saliency kind of trajectory saliency. Results are collected in
Table~\ref{tab:eval_tII} for our three method variants $V\beta_5$,
$V\beta_3$ and $V\beta_0$, the four existing methods, and the baseline.}

The decision threshold $\lambda$ for our method is set on a
validation set with low saliency degree and a saliency ratio of 15\%
since more salient trajectories are expected in this experiment.

{ Actually, the ET-saliency experiment has quite specific characteristics. The salient trajectories are visually different, but not that much. More specifically, they are structurally of the same vein as they share the same entry and exit gates with normal trajectories of a given entry-exit pair. Certainly, ET-saliency is a challenging experiment, since the saliency degree is low. However, it is a specific case study in the sense that essentially the trajectory length counts.

All three variants of our method perform almost equally in terms of F-measure, as shown in Table \ref{tab:eval_tII}. We could expect it, since the key characteristic of the trajectory saliency is here trajectory length. Nevertheless, $V\beta_5$ outperforms $V\beta_0$ regarding the precision score. Our method outperforms the existing methods except for the AET method. Not that surprisingly, the baseline, which is based on the trajectory length, also offers the same performance level (with a threshold set at 10\%), or even outperforms the other methods (with a threshold set at 15\%), apart from the AET method.  However, it is no longer the case in other trajectory-saliency situations as highlighted in Tables \ref{tab:res_gare} and \ref{tab:rs_gamma_all} dealing with DT-saliency, where the baseline performance is lower by a large margin.}

\begin{table}[!htbp]
  \centering
  \begin{tabular}{|c||c|c|c|}
    \hline
    Method  & Precision & Recall & F-measure\\
    variant &           &        &                  \\ \hline
    $V\beta_5$ & 0.92      & 0.67   & 0.77         \\ \hline
    $V\beta_3$ & 0.76      & 0.72   & 0.74     \\ \hline
    $V\beta_0$ & 0.82      & 0.78   & 0.80     \\ \hline
    DAE \cite{Roy2018} &  \textbf{0.95}   & 0.25  &  0.39  \\ \hline
    ALREC \cite{Roy2019}   &  0.90   & 0.33  & 0.48   \\ \hline
    TCDRL-       & & & \\
    KMeans \cite{Yao2017}     &  0.32   & 0.50  &  0.39  \\ \hline
    TCDRL-C \cite{Yao2017}       &  0.73   & 0.61  &  0.67  \\ \hline
    AET \cite{Ma2018}    &  \textbf{0.95}   &  \textbf{1.0}  &  \textbf{0.97}  \\ \hline
    Baseline 10\%      &   0.69   & \textbf{1.0}  &  0.82  \\ \hline
    Baseline 15\%       &  \textbf{0.95}   &  \textbf{1.0} &  \textbf{0.97}  \\ \hline
  \end{tabular}
  \caption{Comparative results on the evaluation set RSTE-CP for the ET-saliency experiment. Precision, recall and F-measure are computed w.r.t. the salient trajectory class. The best performance is in bold.}
  \label{tab:eval_tII}
\end{table}

In addition to this objective experimental comparison, we now discuss inherent differences between our method and existing methods. DAE, ALREC and AET methods build a model to represent the normal data, and then, expect that this model will fail for abnormal elements. The reconstruction error gives the trajectory saliency indicator. A major limitation of this paradigm is its \textit {a priori} low generalisation capability. Indeed, for a new configuration, the normal and abnormal trajectories are likely to change, requiring to train again the network. For a deployment to many different scenarios, this can quickly become unpractical. This approach is suited to detect abnormal trajectories, but less efficient for salient ones. Indeed, abnormal trajectory is an absolute status, meaning it is abnormal in itself, while salient trajectory is a relative status, meaning that it is salient for a given context. If the context changes, the same element may be non salient any more. The AET method performs much better than DAE and ALREC methods, by training a different RNN for each normal trajectory of the training set and by introducing a relative reconstruction error. This gives the AET method greater adaptability, but at the cost of a large number of RNNs to manage and to train.

In contrast to this family of methods, our approach is able to handle relative saliency, as illustrated by the experiments presented in Sections~\ref{subsec:syn} and \ref{subsec:res_gare}. Indeed, for these experiments, a trajectory appears as salient in a given context, while
in a different one the same trajectory may be non
salient. Furthermore, we do not rely on the assumption that a salient trajectory is not present in the training set. In fact, if salient trajectories are included in the training set, we expect they will be properly reconstructed and encoded, thus facilitating the decision. It enables us to train the network only once for the dataset, whatever the scenario considered.~\\

\begin{table}[!htbp]
  \centering
  \begin{tabular}{|c||c|c|c|}
    \hline
    Method  & Precision & Recall & F-measure\\
    variant &           &        &                  \\ \hline
    $V\beta_5$ & 0.82      & 0.98   & 0.89         \\ \hline
    $V\beta_3$ & 0.87      & \textbf{1.0}    & \textbf{0.93}     \\ \hline
    $V\beta_0$ & 0.88      & 0.99   & \textbf{0.93}     \\ \hline
    AET        & \textbf{0.92}      & 0.86   & 0.89     \\ \hline
  \end{tabular}
  \caption{Results on the evaluation set RTSE-LS with FT-saliency, obtained with our three metod variants and with the AET method. Precision, recall and F-measure are computed w.r.t. the salient trajectory class. The best performance is in bold.}
  \label{rste-LS-FT}
\end{table}

\subsubsection{Results on the evaluation set RSTE-LS with FT-saliency}

In this section, we present results obtained on a large scale using the evaluation set RSTE-LS with the FT-saliency kind of trajectory saliency.
Let us recall that for the FT-saliency kind, salient trajectories are trajectories of the very same scenario but with a faster velocity. For example, they might correspond to few people who run, as opposed to the vast majority of people who walk in the station corridors. However, we needed to simulate such trajectories to carry out a relevant FT-saliency experiment. We created the salient trajectories in each scenario (i.e., each entry-exit pair) by sub-sampling a small amount of trajectories of the scenario. In practice, the sub-sampling rate was set to 3, meaning that the salient trajectories were three-times faster. The ratio of salient trajectories is set to 5\%.

{ Results obtained by our three method variants are reported in Table \ref{rste-LS-FT}. We also involved the AET method in this FT-saliency experiment. We observe that our method provides overall very good detection rates, even if the saliency degree is low, the difference being only in the path velocity. The F-measure scores of all tested methods are relatively close, certainly because normal and salient trajectories keep the same global shape in the FT-saliency experiment. Nevertheless, the best results are provided by our two method variants $V\beta_0$ and $V\beta_3$ that perform better than AET method.}

\begin{table}[!htbp]
  \centering
  \begin{tabular}{|c|c||c|c|c|}
    \hline
    Method  & Saliency   & Precision & Recall & F-measure\\
    variant & degree      &          &        &        \\ \hline
    $V\beta_5$ & Low     & \textbf{0.65}     & \textbf{0.73}    & \textbf{0.69}       \\ \hline
    $V\beta_3$ & Low     & 0.49     & 0.56    & 0.52     \\ \hline
    $V\beta_0$ & Low     & 0.53     & 0.59    & 0.56     \\ \hline
    AET \cite{Ma2018} & Low     & 0.58     & 0.45    & 0.51     \\ \hline
    \hline
    $V\beta_5$ & Medium & \textbf{0.88}      & \textbf{0.99}   & \textbf{0.93}     \\ \hline
    $V\beta_3$ & Medium & 0.73      & 0.89   & 0.80    \\ \hline
    $V\beta_0$&  Medium & 0.82      & 0.89   & 0.85     \\ \hline
    AET \cite{Ma2018} &  Medium & 0.82      & 0.65   & 0.72     \\ \hline
    \hline
    $V\beta_5$ & High   & \textbf{1.0}      & \textbf{1.0}   & \textbf{1.0}      \\ \hline
    $V\beta_3$ & High   & 0.97     & 0.99   & 0.98       \\ \hline
    $V\beta_0$ & High   & 0.95     & \textbf{1.0}   & 0.97       \\ \hline
    AET \cite{Ma2018} & High   & 0.96     & 0.91   & 0.93       \\ \hline
  \end{tabular}
  \caption{Results on the evaluation set RSTE-LS for DT-saliency obtained with our three method variants for a saliency ratio of 5\%. Threshold $\lambda$ was set to 2 for all the experiments. For each saliency degree, the best performance is in bold. The highest the saliency degree, the easiest the case. Results obtained with the AET method are also included.}
  \label{tab:rs_gamma_all}
\end{table}
\subsubsection{Results on the evaluation set RSTE-LS with DT-saliency}
{ We now report results obtained for the DT-saliency experiment on the RSTE-LS evaluation set.}
Table~\ref{tab:rs_gamma_all} contains the results obtained for the three variants of our method, $V\beta_5$, $V\beta_3$ and $V\beta_0$, for different degrees of saliency. Let us recall that $V\beta_5$, $V\beta_3$ and $V\beta_0$ are the variants for which the parameter on the consistency constraint is respectively $10^5$, $10^3$ and $0$. For all these experiments, we set the threshold $\lambda$ to a given value ($\lambda=2$) once and for all. First of all, we observe as expected that the more salient trajectories depart from normal ones, the better the saliency
estimation results are. { Then, among our method variants, $V\beta_5$, $V\beta_3$ and $V\beta_0$, the best method for a given degree of saliency is $V\beta_5$, and the more challenging the experiment (i.e., the lowest the saliency degree), the larger the margin. This confirms the interest of the consistency constraint. We also include results obtained with the AET method, since it provided very good results on RSTE-CP. For this experiment, we trained about 900 RNNs for the AET method. All variants of our method outperform the AET method in this experiment that is more challenging and representative than the ones on the RSTE-CP evaluation set. Indeed, this experiment is on a large scale, and the same trajectory may be salient or normal depending on the context (i.e., the scenario). Moreover, for the more difficult cases (i.e., the low and medium saliency degrees), they outperform it by a large margin.}

{ Finally, we carried out a complementary DT-saliency experiment} that consists in varying the ratio of saliency in the test dataset (defined as the ratio between the number of salient and normal trajectories). Results are collected in
Table~\ref{tab:rs_gamma_var_ratio} for the best variant $V\beta_5$. We
observe that the overall performance evaluated with the F-measure does
not vary much when the ratio of saliency increases, which demonstrates
that our method is applicable for different saliency regimes.

\begin{table}[!htbp]
  \centering
  \begin{tabular}{|c|c|c||c|c|c|}
    \hline
    Method  & Saliency   & Saliency & Precision & Recall & F-measure\\
    variant & degree & ratio    &           &        &          \\ \hline
    $V\beta_5$& Low        & 5\%      & 0.65     & 0.73  & 0.69    \\ \hline
    $V\beta_5$& Low        & 10\%     & 0.85     & 0.61  & 0.71    \\ \hline
    $V\beta_5$& Low        & 15\%     & 0.91     & 0.51  & 0.66    \\ \hline
 \hline
    $V\beta_5$& Medium     & 5\%      & 0.88     & 0.99  & 0.93    \\ \hline
    $V\beta_5$& Medium     & 10\%     & 0.96     & 0.96  & 0.96    \\ \hline
    $V\beta_5$& Medium     & 15\%     & 1.0      & 0.89  & 0.94    \\ \hline
    \hline
    $V\beta_5$& High       & 5\%      & 1.0      & 1.0   & 1.0     \\ \hline
    $V\beta_5$& High       & 10\%     & 1.0      & 0.99  & 1.0     \\ \hline
    $V\beta_5$& High       & 15\%     & 1.0      & 0.91  & 0.96    \\ \hline

  \end{tabular}
  \caption{Results on the evaluation set RSTE-LS with DT-saliency, for different saliency ratios, obtained with the best performing variant of our method. Threshold $\lambda$ was set to 2 for all the experiments.}
  \label{tab:rs_gamma_var_ratio}
\end{table}

\begin{figure}[!htbp]
  \centering
  \begin{tabular}[!htbp]{c}
    \includegraphics[width=0.8\linewidth]{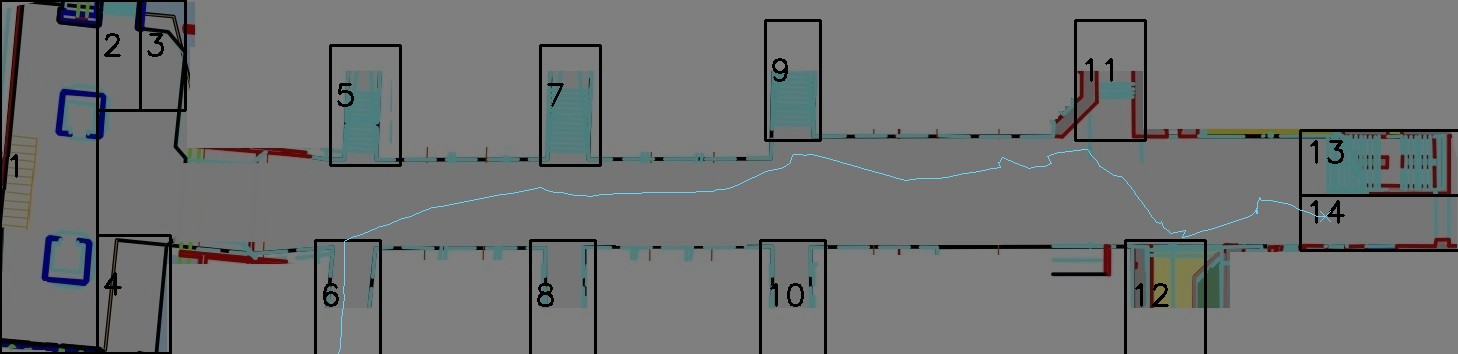}   \\
    \includegraphics[width=0.8\linewidth]{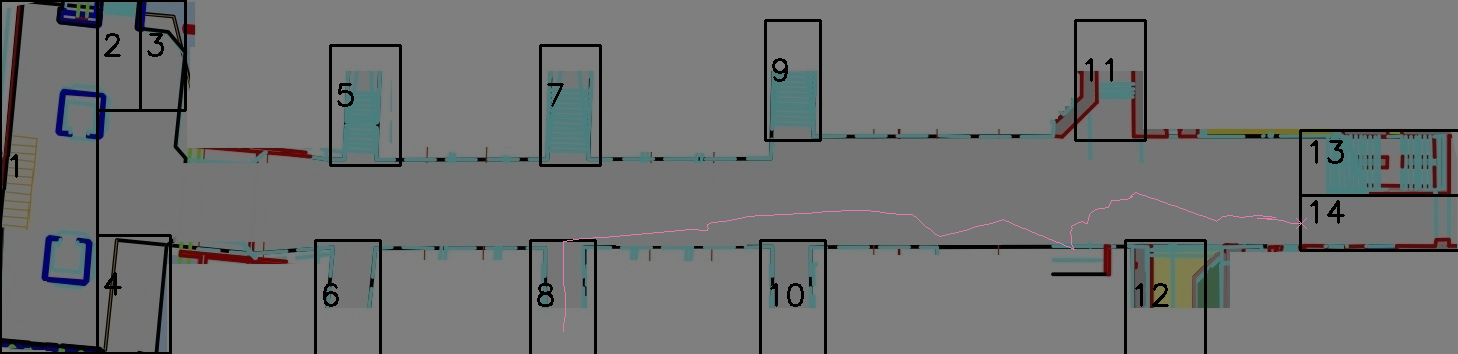}   \\
    \includegraphics[width=0.8\linewidth]{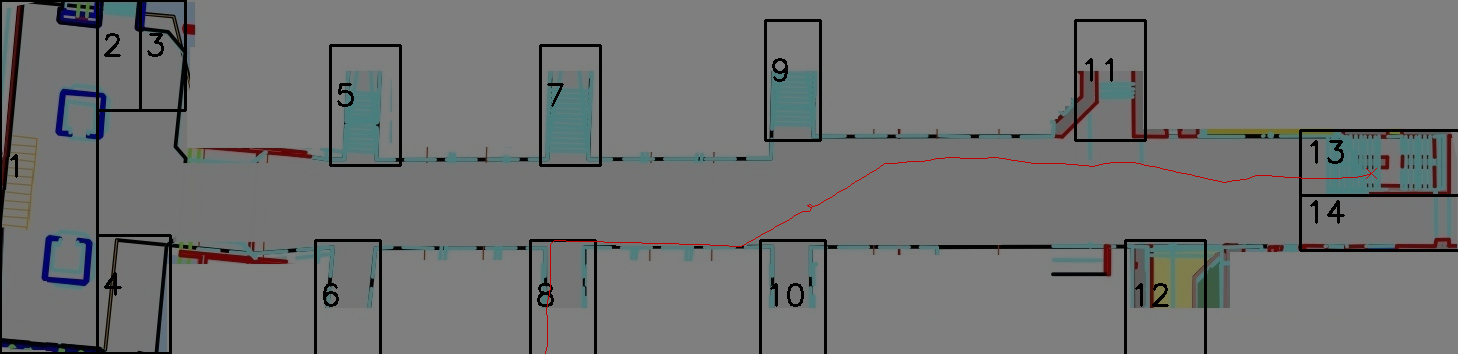}   \\
    \includegraphics[width=0.8\linewidth]{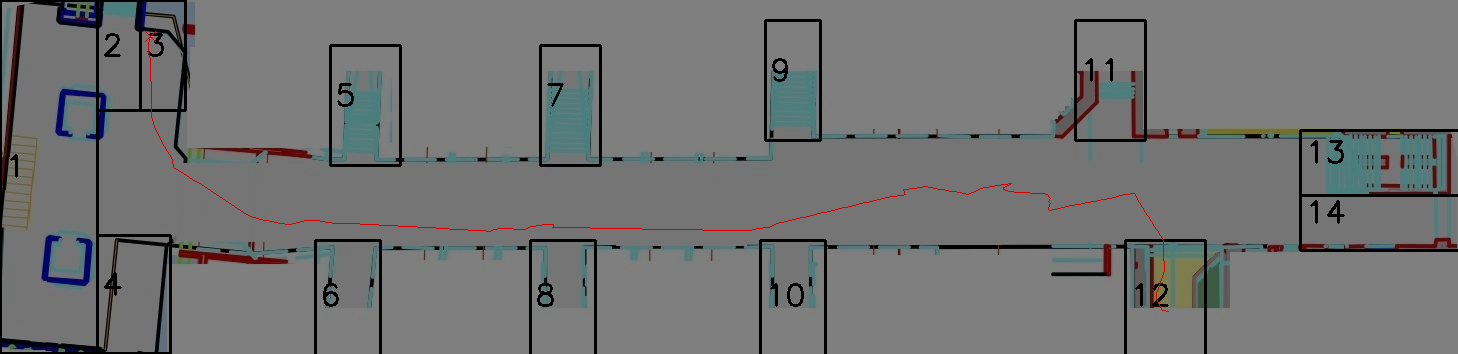}   \\
  \end{tabular}
  \caption{Four failure cases for RSTE-LS with DT-saliency, provided by the method $V\beta_5$, for a low saliency degree and a saliency ratio of 5\%. The two top trajectories are not salient but are predicted as salient. The two bottom trajectories are salient but have been classified as normal.}
  \label{fig:failure_cases}
\end{figure}

In addition, we show four representative failure cases for the method $V\beta_5$ in Figure~\ref{fig:failure_cases}. They were obtained in the RSTE-LS test set with DT-saliency, for a low saliency degree and a saliency ratio of 5\%. The two top trajectories are not salient but are predicted as salient. We can observe that these two trajectories present minor irregularities. The two bottom trajectories are salient but have been classified as normal. For the first one, the normal trajectories of this scenario start at gate 14, which is very close to gate 13, the starting point of this salient trajectory. For the second one, the length of the trajectory may make the prediction harder, since normal trajectories go from gate 3 to 14 in this scenario. In general, failure cases occur when the margin of decision was small as in these four examples.

\subsection{Additional investigations and visualisations}
\label{sec:cmpl_exp}

We now present additional investigations about the training, latent
code estimation and decision stages. They will enable to better
understand the behaviour of our method, and to provide more details on
a few specific issues.

\subsubsection{Initial state of the LSTMs}

The network designed for trajectory representation includes LSTM
layers in the encoder and the decoder. In a recurrent network, the
value predicted for the next time step is estimated with the help of a
hidden state that keeps memory of past information. However, there is no past information for the first time step. In practice, there are several ways to initialise the hidden LSTM state (see for instance \cite{Zimmermann2012}). The simplest solution consists in setting the initial state to a constant value, by default to 0. Slightly more sophisticated possibilities consist in setting it randomly or in learning it. If the initial value is randomly chosen, it is a way to introduce data augmentation, since the network faces more diverse configurations for the same training dataset. It may consequently help improving performance. Learning the initial state can be done by including the initial state as a variable in the backpropagation algorithm.  In preliminary experiments, we did not find any significant difference between these different initialisations in the network performance. We then decided to adopt a random initialisation, by drawing the initial state from a normal distribution.

\begin{figure}[!htbp]
  \centering
  \begin{tabular}{c}
    \includegraphics[width=0.8\linewidth, height=2.6cm]{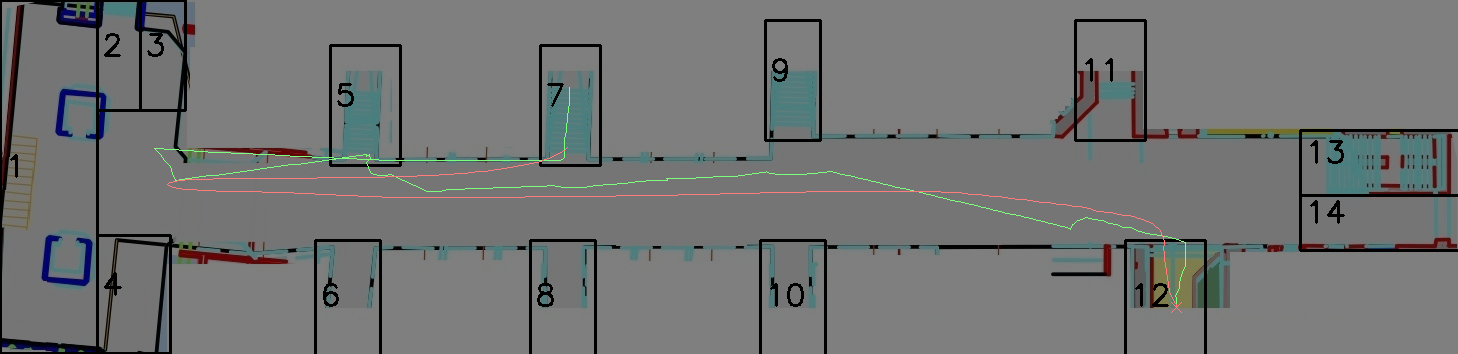}   \\
    \includegraphics[width=0.8\linewidth, height=2.6cm]{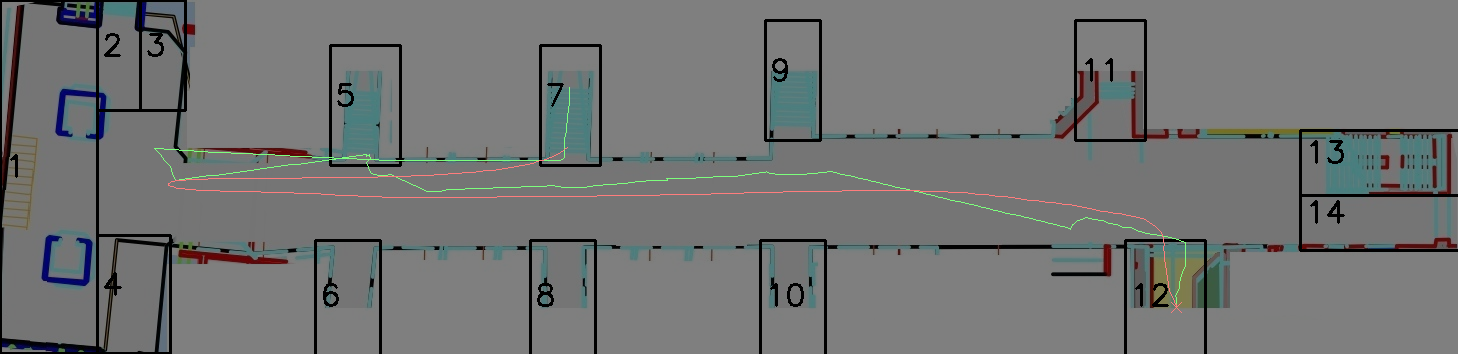}   \\
    \includegraphics[width=0.8\linewidth, height=2.6cm]{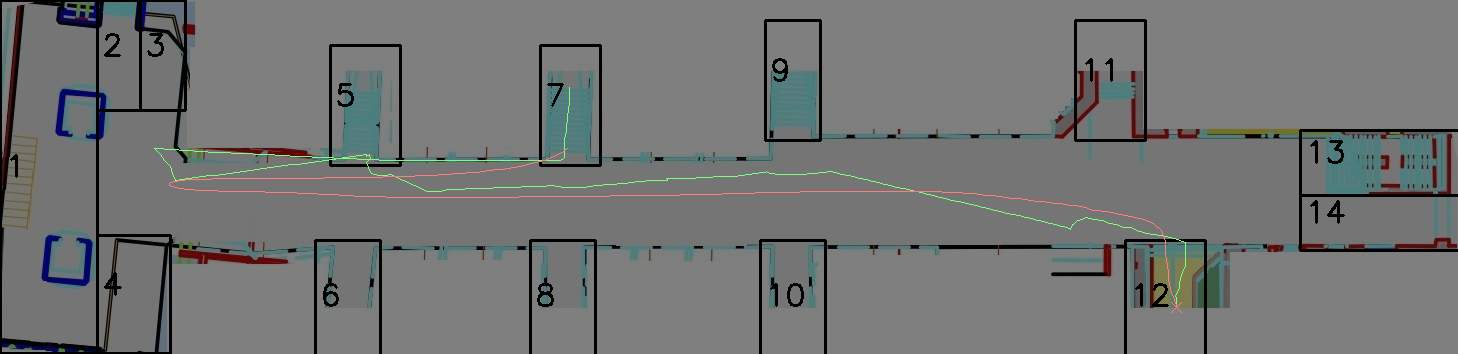}   \\
  \end{tabular}
  \caption{One trajectory reconstructed three times by our network
    with different random initialisations of the hidden state of the
    LSTM. The reconstructed trajectory is drawn in red and the ground
    truth trajectory is drawn in green. The impact on the reconstructed trajectory is minimal.}
  \label{fig:lstm_rand_init}
\end{figure}

With a random initialisation of the hidden state, the resulting
parametrisation of the network is of course not strictly
deterministic. Still, the network delivers consistent predictions when the same trajectory is given as input several times. This is
illustrated in Figure~\ref{fig:lstm_rand_init} displaying several
reconstructions of the same trajectory for several random
initialisations. Apart from a hardly visible variation near the
starting position of the trajectory (bottom-right), the
reconstructions are almost identical.

\subsubsection{Evolution of the model during training}

Our method builds constrained latent codes to represent trajectories, and the codes are exploited to detect trajectory saliency. Our decision framework takes benefit from the code position in the latent space. It is somewhat inspired by deep metric learning.

\begin{figure}[!htbp]
  \centering
  \begin{tabular}{cc}
    \includegraphics[width=0.45\linewidth]{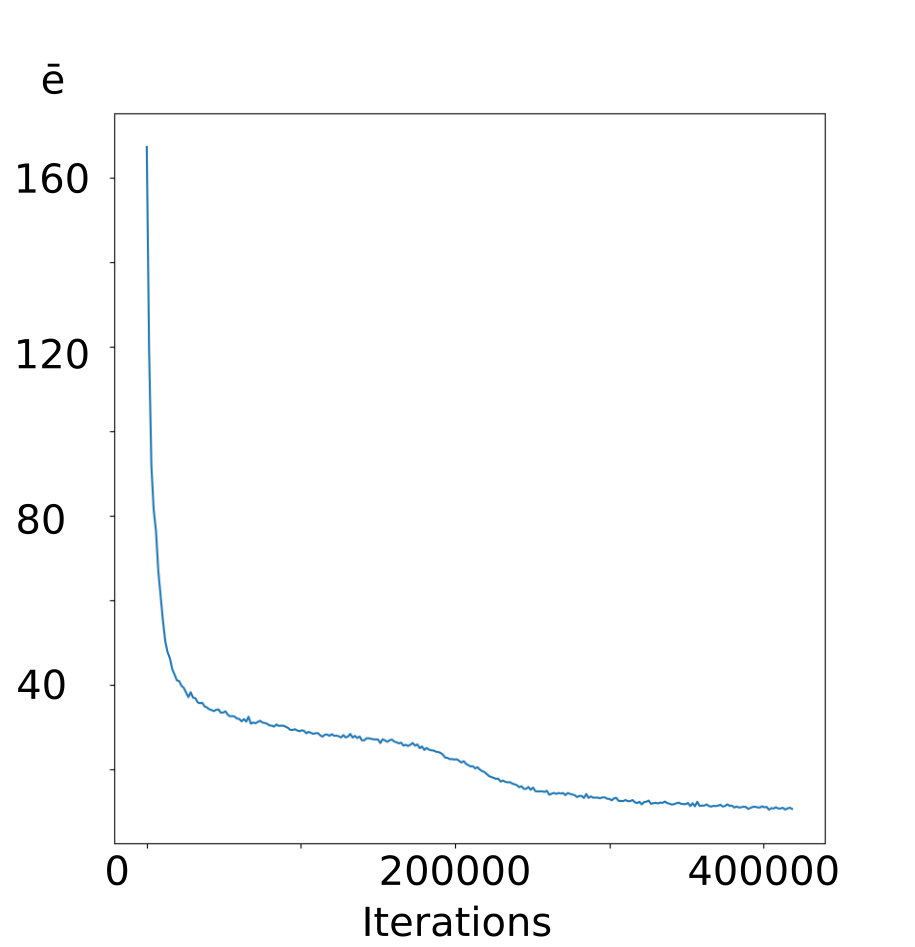}
    & \includegraphics[width=0.45\linewidth]{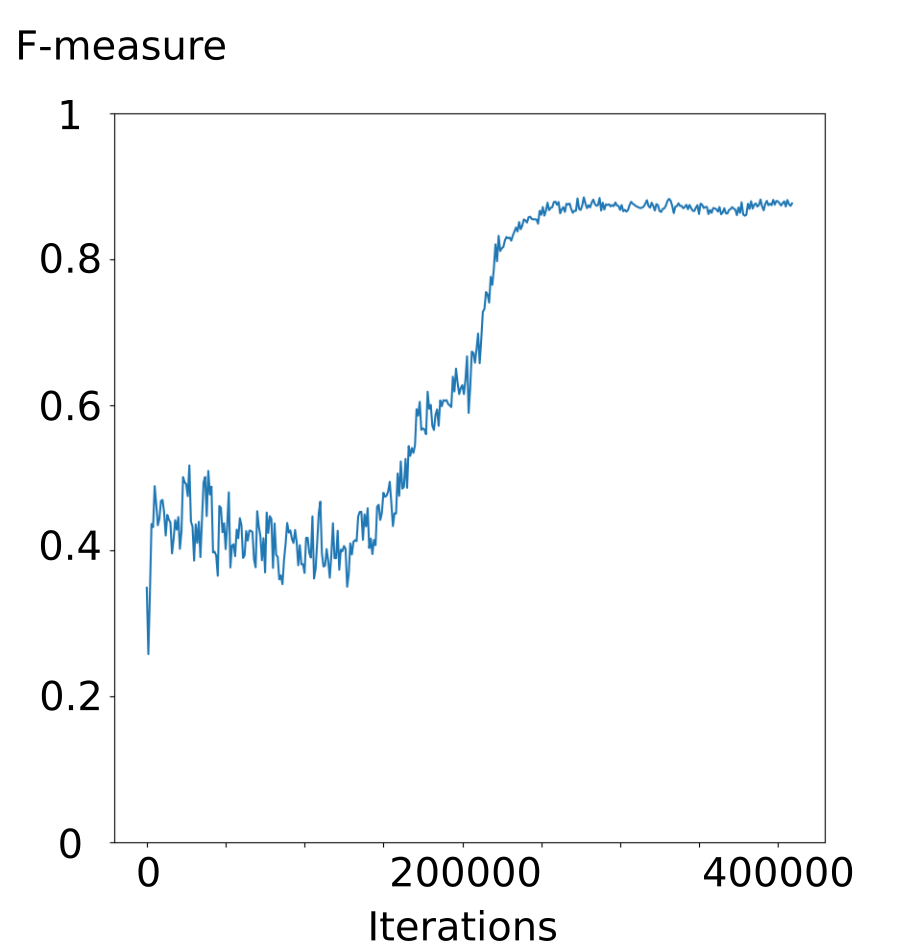}
  \end{tabular}
  \caption{Evolution of the trajectory reconstruction error $\bar{e}$
    (left) and of the F-measure related to trajectory saliency
    detection (right) on the validation set of the STMS dataset during
    training. Each iteration corresponds to a batch of 60 to 66
    trajectories (depending on the random inclusion of salient
    trajectories).}
  \label{fig:train_crv}
\end{figure}

Figure~\ref{fig:train_crv} displays the evolution of the trajectory
reconstruction error and the F-measure related to trajectory saliency detection on the STMS validation set along the training
iterations. Each iteration corresponds to a batch of 60 to 66
trajectories, depending on the random inclusion of salient
trajectories. We observe that, after an initial phase with large
improvements, the gain in performance becomes slower and slower. There is a noticeable improvement around iteration 200,000, even if the final gain remains limited. On the other hand, regarding saliency detection, performance reaches a plateau in a few iterations, and even deteriorates a little for a while. Only after more than 100,000 iterations, performance improves again relatively quickly, before reaching another plateau.

A similar behaviour is observed for deep metric learning
\cite{Hermans2017}. Indeed, codes are expected to evolve in the latent space during learning, in order to correctly represent different elements into separate clusters. However, as long as the clusters are not clearly distinguishable, the saliency detection algorithm cannot work well. As a consequence, it is beneficial to let the training go on, even if saliency detection performance apparently stagnates.

\subsubsection{Code distribution}

The consistency constraint was designed to make codes representing
similar elements closer. To better understand the effects of this
constraint, we computed the empirical distribution of the learned
codes with and without the consistency constraint on the STMS
validation dataset. The components of each code are plotted as
histograms in Table~\ref{tab:codes_vis}. The components corresponding
to a training without consistency are denoted $f_i$, and those with
consistency are denoted $g_i$. Their values lie in $[-1, 1]$.

Without the consistency constraint, all the code components vary
largely. Two main patterns stand out. Either the code values are
spread in [-1,1], or the code values are restricted to positive or
negative ones. When the consistency constraint is applied, the
variability is far lower. In almost all cases, the difference between
the smallest and largest values is smaller than 0.5, and for several
components, the predicted value is practically constant.

The consistency constraint has then a clear impact on the
codes. Without it, the network tends to take profit of all the
possible values to reconstruct the trajectories with high
precision. When enforcing the constraint, we end up with far smaller
variations and only the main significant information is stored. Local
variations inherent to each trajectory are more likely to be
discarded.

\begin{table}[!htbp]
  \begin{tabular}{c||c}
    \hline
    \begin{tabular}{cc}
      \includegraphics[width=0.17\linewidth,height=0.17\linewidth, trim=0 0 0 -20]{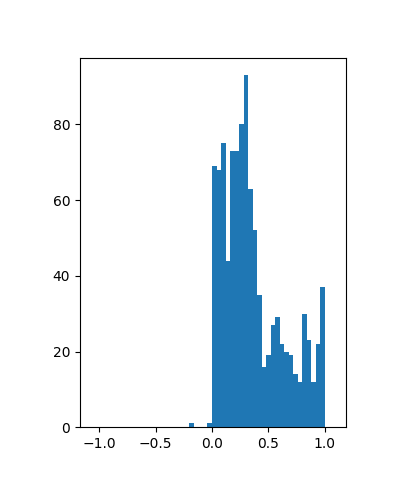}
      &\includegraphics[width=0.17\linewidth,height=0.17\linewidth, trim=0 0 0 -20]{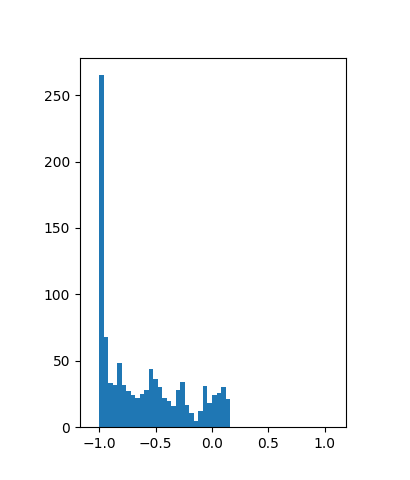} \\
      $f_{0}$ & $f_{1}$
    \end{tabular}
      & \begin{tabular}{cc}
       \includegraphics[width=0.17\linewidth,height=0.17\linewidth, trim=0 0 0 -20]{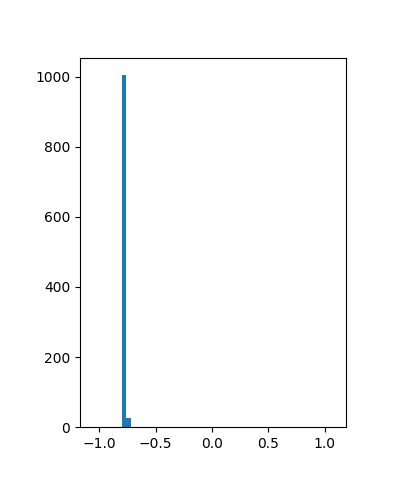}
       &\includegraphics[width=0.17\linewidth,height=0.17\linewidth, trim=0 0 0 -20]{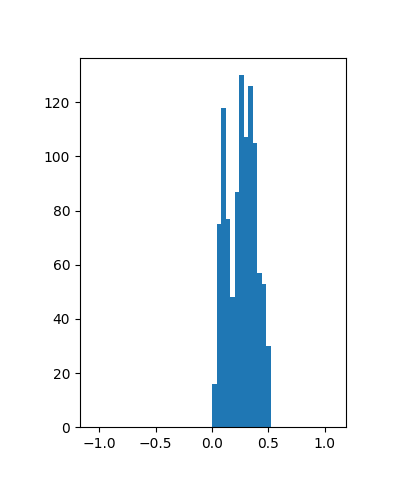} \\
       $g_{0}$ & $g_{1}$ \\
    \end{tabular} \\ \hline
    \begin{tabular}{cc}
    \includegraphics[width=0.17\linewidth,height=0.17\linewidth, trim=0 0 0 -20]{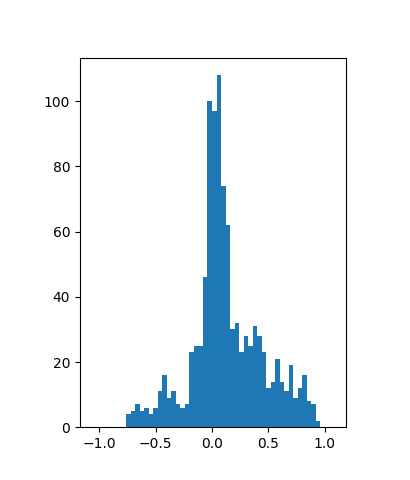}
    &\includegraphics[width=0.17\linewidth,height=0.17\linewidth, trim=0 0 0 -20]{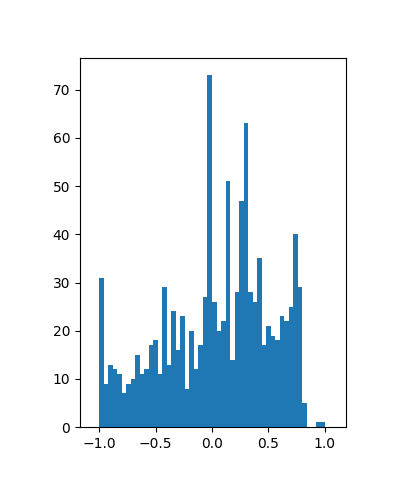} \\
    $f_{2}$ & $f_{3}$
    \end{tabular}
    & \begin{tabular}{cc}
       \includegraphics[width=0.17\linewidth,height=0.17\linewidth, trim=0 0 0 -20]{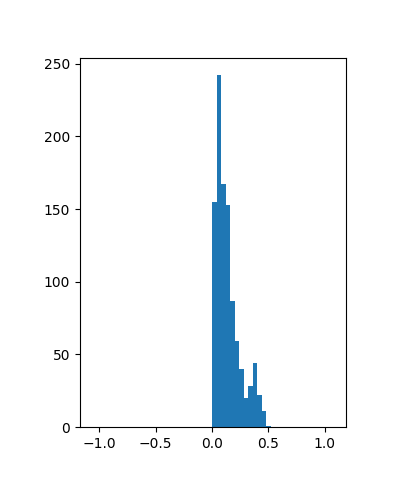}
       &\includegraphics[width=0.17\linewidth,height=0.17\linewidth, trim=0 0 0 -20]{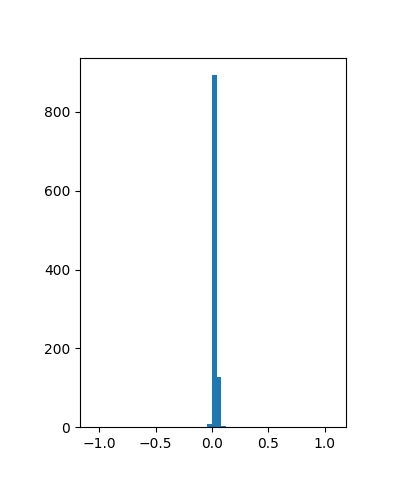} \\
       $g_{2}$ & $g_{3}$ \\
    \end{tabular} \\ \hline
    \begin{tabular}{cc}
    \includegraphics[width=0.17\linewidth,height=0.17\linewidth, trim=0 0 0 -20]{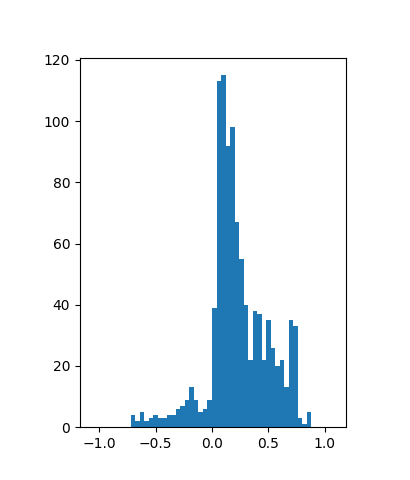}
    &\includegraphics[width=0.17\linewidth,height=0.17\linewidth, trim=0 0 0 -20]{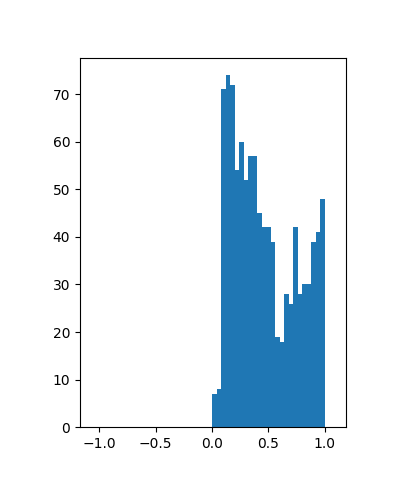} \\
    $f_{4}$ & $f_{5}$
    \end{tabular}
    & \begin{tabular}{cc}
       \includegraphics[width=0.17\linewidth,height=0.17\linewidth, trim=0 0 0 -20]{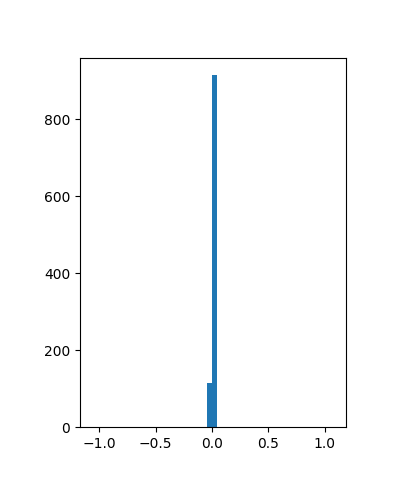}
       &\includegraphics[width=0.17\linewidth,height=0.17\linewidth, trim=0 0 0 -20]{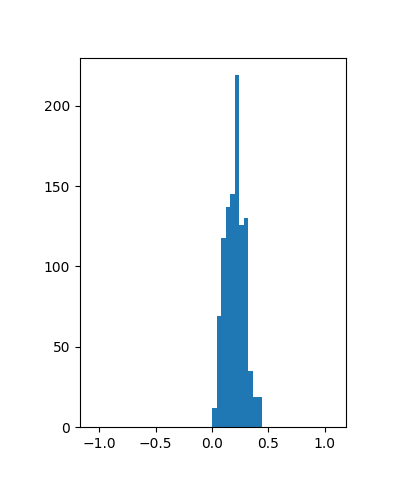} \\
       $g_{4}$ & $g_{5}$ \\
    \end{tabular} \\ \hline
    \begin{tabular}{cc}
    \includegraphics[width=0.17\linewidth,height=0.17\linewidth, trim=0 0 0 -20]{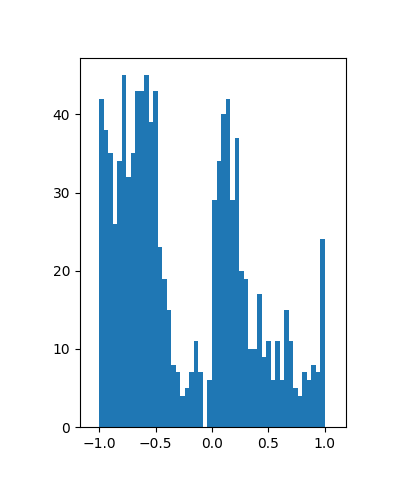}
    &\includegraphics[width=0.17\linewidth,height=0.17\linewidth, trim=0 0 0 -20]{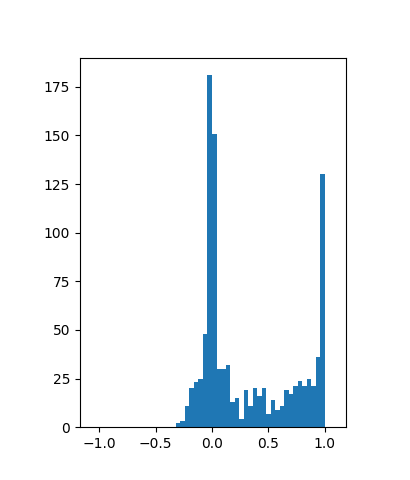} \\
    $f_{6}$ & $f_{7}$
    \end{tabular}
    & \begin{tabular}{cc}
       \includegraphics[width=0.17\linewidth,height=0.17\linewidth, trim=0 0 0 -20]{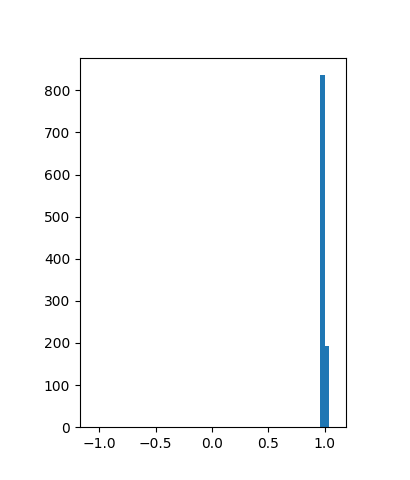}
       &\includegraphics[width=0.17\linewidth,height=0.17\linewidth, trim=0 0 0 -20]{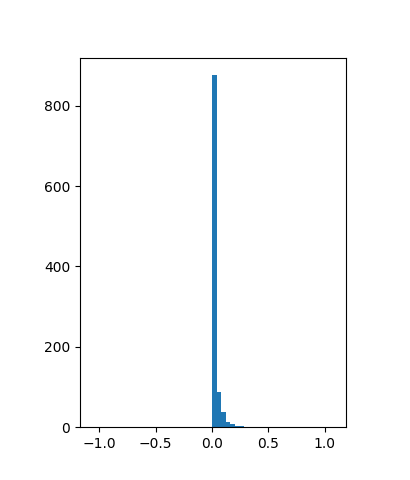} \\
       $g_{6}$ & $g_{7}$ \\
    \end{tabular} \\ \hline
    \begin{tabular}{cc}
    \includegraphics[width=0.17\linewidth,height=0.17\linewidth, trim=0 0 0 -20]{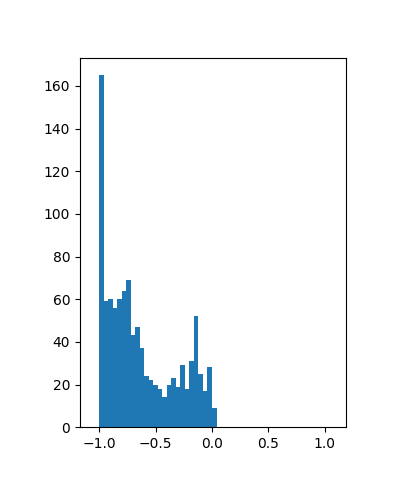}
    &\includegraphics[width=0.17\linewidth,height=0.17\linewidth, trim=0 0 0 -20]{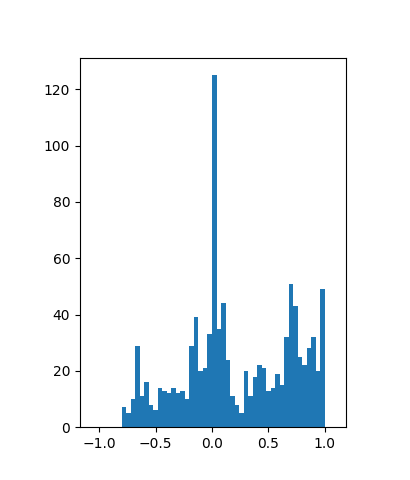} \\
    $f_{8}$ & $f_{9}$
    \end{tabular}
    & \begin{tabular}{cc}
       \includegraphics[width=0.17\linewidth,height=0.17\linewidth, trim=0 0 0 -20]{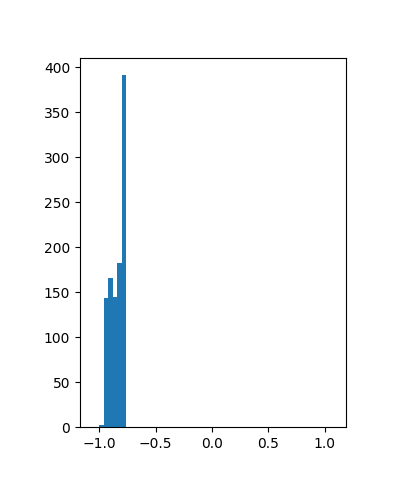}
       &\includegraphics[width=0.17\linewidth,height=0.17\linewidth, trim=0 0 0 -20]{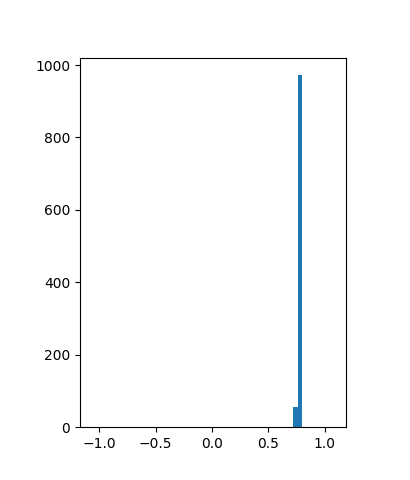} \\
       $g_{8}$ & $g_{9}$ \\
    \end{tabular} \\ \hline
    \begin{tabular}{cc}
    \includegraphics[width=0.17\linewidth,height=0.17\linewidth, trim=0 0 0 -20]{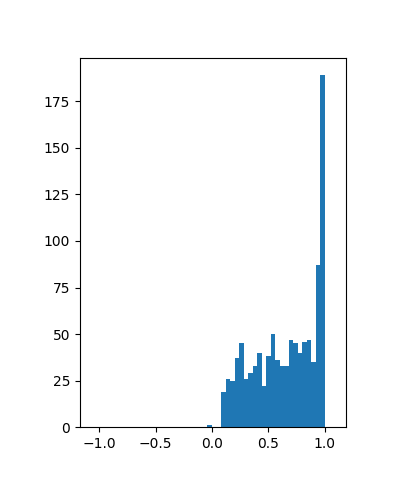}
    &\includegraphics[width=0.17\linewidth,height=0.17\linewidth, trim=0 0 0 -20]{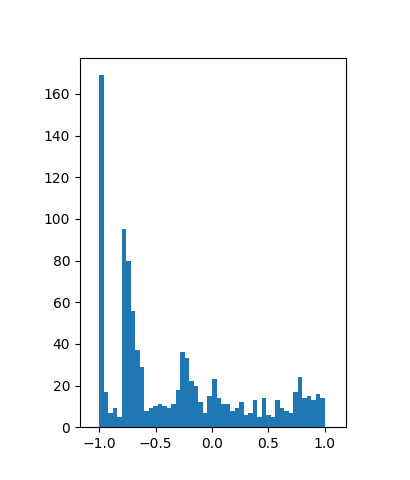} \\
    $f_{10}$ & $f_{11}$
    \end{tabular}
    & \begin{tabular}{cc}
       \includegraphics[width=0.17\linewidth,height=0.17\linewidth, trim=0 0 0 -20]{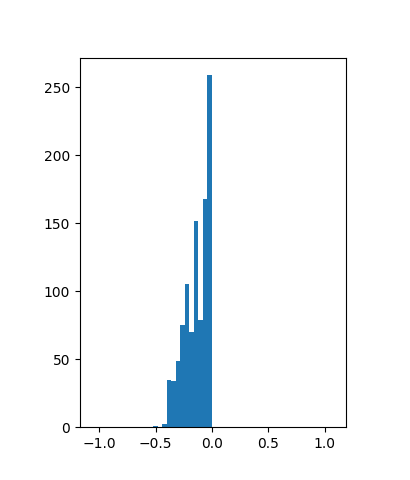}
       &\includegraphics[width=0.17\linewidth,height=0.17\linewidth, trim=0 0 0 -20]{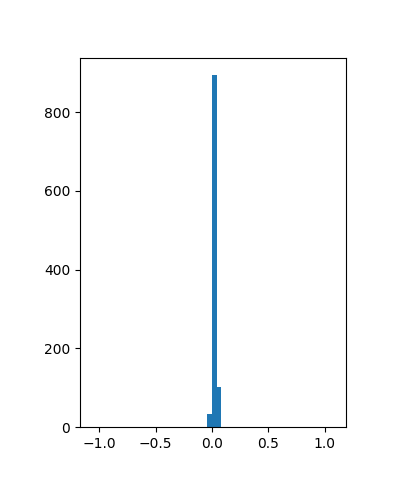} \\
       $g_{10}$ & $g_{11}$ \\
    \end{tabular} \\ \hline
  \end{tabular}
  \caption{Histograms of the first components of the trajectory codes,
    after training on STMS dataset. The remaining components exhibit a
    similar behaviour. The $f_i$ components on the left correspond to
    a training without the consistency constraint, and the $g_i$ on
    the right correspond to a training with the consistency
    constraint. The dispersion of the components is far more limited
    with the consistency constraint as expected.  }
  \label{tab:codes_vis}
\end{table}

\subsubsection{Influence of choice of $\lambda$}

\begin{figure}[!htbp]
  \centering
  \includegraphics[width=0.84\linewidth]{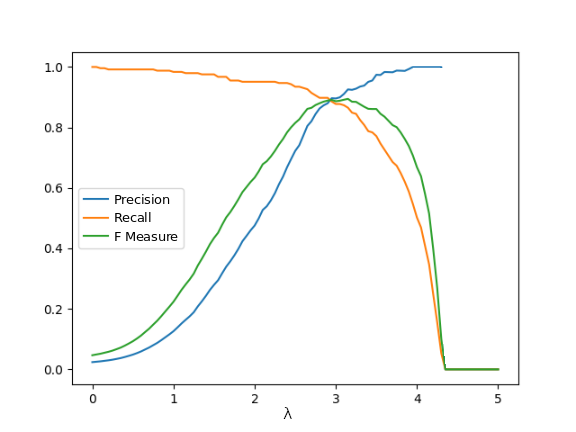}
  \caption{Precision, recall and F-measure computed w.r.t. the salient
    trajectory class. These metrics quantify the trajectory saliency
    detection performance for different values of $\lambda$ on the
    STMS validation set. $\lambda$ values are sampled every 0.05
    within [0,5].}
  \label{fig:sal_lambda}
\end{figure}

To assess the influence of the threshold value on the method
performance, the trajectory saliency detection was conducted on the
STMS validation set with a regular sampling of $\lambda$ (every 0.05)
in the interval $[0, 5]$. The corresponding precision, recall and
F-measures are plotted in Figure~\ref{fig:sal_lambda}. It shows that
in this case there is a plateau around $\lambda = 3$, which means that
the performance is not sensitive to the decision threshold value. This
observation is consistent with the results reported in
Tables~\ref{tab:rs_gamma_all} and \ref{tab:rs_gamma_var_ratio} on the
real dataset RTS, where the same threshold value was used for all the
experiments. The threshold value is of course dependent on the dataset
and the nature of the trajectories.

\section{Conclusion}
\label{sec:concl}

We have defined a novel framework for trajectory saliency detection
based on a recurrent auto-encoder supplying a compact latent
representation of trajectories. The auto-encoder training includes a
consistency constraint in the loss function. The proposed method
remains almost unsupervised, while being able to find saliency even
when the difference with normality is subtle. We have explicitly
formulated the trajectory saliency state and its associated decision
algorithm. A trajectory is not salient in itself and so is not
labelled only because a similar occurrence has never been seen in the training data. Trajectory saliency is defined with respect to a given context that specifies the normal trajectories. Such a paradigm allows for an easy generalisation to new configurations. Indeed, a major advantage of our method is its flexibility and its essentially unsupervised nature. We have experimentally validated our trajectory saliency detection method and its main components on synthetic and real trajectory datasets. We have demonstrated on several trajectory-saliency experiments drawn from a publicly available dataset of pedestrian trajectories that our method compares favourably to existing methods.

\appendix

\section{\textit{p}-value method to set the saliency threshold}
\label{sec:alt_lambda}

We report experiments on the \textit{p}-value scheme to set the
$\lambda$ threshold for the trajectory saliency detection. We tested
three probability distributions, the Weibull distribution
\cite{Weibull1939}, the Dagum distribution \cite{Dagum1977} in the
standardised form and the Dagum distribution in the general form, to
fit the empirical distribution of the $q_i$ descriptors defined in
eq.(\ref{eq:quo}).

\begin{figure}[!htbp]
  \centering
  \includegraphics[width=0.84\linewidth]{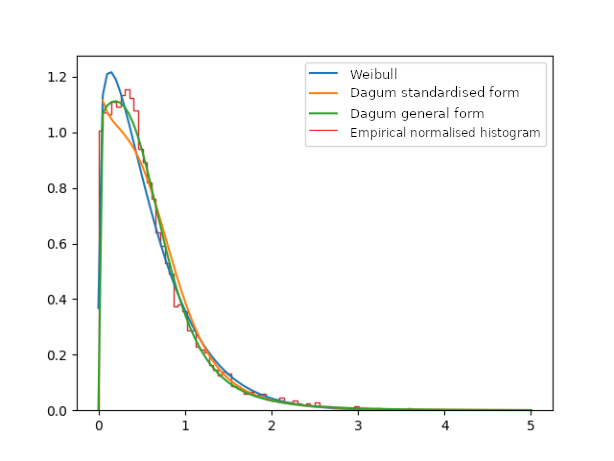}
  \caption{Plots of the empirical distribution of the $q_i$
    descriptors and the fitted distributions. They are computed from
    the 10,000 normal trajectories of the STMS validation dataset. Each scenario is processed separately to compute the $q_i$
    descriptors.}
  \label{fig:fit_proba}
\end{figure}

Empirical and estimated probability distributions are plotted in
Figure~\ref{fig:fit_proba} for the STMS validation dataset. From this set, 10,000 $q_i$'s corresponding to normal trajectories are
computed. Distributions are fitted with the maximum likelihood method.

Visually, the best fit is obtained with the general form of the Dagum distribution. In addition, to get a quantitative measure of the fitting error, we use the criterion $\mathcal{F}$ defined as follows:
\begin{equation}
  \label{eq:eval_fitting_law}
  \mathcal{F} = \sum_b | \mathcal{G}(b) - \mathcal{H}(b) |
\end{equation}
with $b$ a bin, $\mathcal{G}(b)$ the area under the curve of the
fitted probability law for the bin $b$, and $\mathcal{H}(b)$ the
observed proportion of elements falling into the bin $b$. In practice, we use 101 bins regularly sampled between 0 and 5. The fitting errors are 0.05, 0.10 and 0.08 respectively for the Dagum distribution with the general form, the Dagum distribution with the standardised form and the Weibull distribution, confirming the visual evaluation. Consequently, we adopted the Dagum distribution with the general form.

Once the three parameters of the general Dagum distribution are
estimated, we can fix the \textit{p}-value and then get the threshold value $\lambda$. In an ideal case, the estimated distribution should correspond also to the test data. At this point, let us emphasise that we have no guarantee that this is the case. Indeed, the $q_i$ descriptors depend on the presence of outliers, that is, salient trajectories, through the normalisation with $\bar{d}$ and $\sigma$, the computed mean and standard deviation. In particular, the standard deviation $\sigma$ may be noticeably influenced by salient trajectories. Indeed, it tends to make the $q_i$ smaller. As a consequence, fewer $q_i$'s will be above the threshold $\lambda$.

The \textit{p}-value approach converts the problem of setting the
value of $\lambda$ into the choice of the tolerated proportion of
false positive in the prediction. In contrast, $\lambda$ is not
directly interpretable. The issue is that at test time, the
\textit{p}-value does not correspond actually to the expected
proportion of false positives. Instead, it only provides an upper
bound for this quantity. If the upper bound is too large, the
\textit{p}-value scheme will be no longer effective.

To test the \textit{p}-value scheme, we took the evaluation setting
RSTE-A with the lowest saliency degree. To fit the distribution, we
need a subset of only normal trajectories. The TrG set meets this
criterion. We used the $V\beta_5$ variant. Results are given in
Table~\ref{tab:rs_pval} for saliency ratios from 0.05 to 0.15. We give the chosen \textit{p}-value and the False Positive Rate (FPR) that corresponds to the ratio between the number of false positives and the number of normal trajectories.

Results confirm that the empirical probability distribution of the
$q_i$ descriptors changes in presence of outliers. Indeed, the
\textit{p}-value and the FPR, which should normally have similar
values (the FPR can be viewed as the empirical \textit{p}-value),
differ largely in practice.

\begin{table}[!htbp]
  \centering
  \begin{tabular}{|c||c|c|c||c|c|c|}
    \hline
    Saliency & \textit{p}-value & $\lambda$ & FPR &  P & R & FM\\
    ratio    &  &  &  &         &        &          \\ \hline \hline

    5\%      & 0.025 & 2.4 & 0.010 & 0.73 & 0.64  & 0.68  \\ \hline
    5\%      & 0.05  & 1.9 & 0.019 & 0.61 & 0.73  & 0.67  \\ \hline
    10\%     & 0.05  & 1.9 & 0.012 & 0.83 & 0.62  & 0.71  \\ \hline
    10\%     & 0.10  & 1.5 & 0.035 & 0.67 & 0.76  & 0.71  \\ \hline
    15\%     & 0.075 & 1.7 & 0.016 & 0.85 & 0.63  & 0.72  \\ \hline
    15\%     & 0.15  & 1.3 & 0.050 & 0.67 & 0.73  & 0.70  \\ \hline

  \end{tabular}
  \caption{Evaluation setting RSTE-A, and $\lambda$ set with the
    \textit{p}-value scheme. P, R and FM denote respectively
    precision, recall and F-measure w.r.t. the salient trajectory
    class. FPR denotes the False Positive Rate.}
  \label{tab:rs_pval}
\end{table}

\section*{Acknowledgement}

\noindent
This work was supported in part by the DGA and the Région Bretagne
through co-funding of Léo Maczyta's PhD thesis.

\bibliographystyle{amsalpha}
\bibliography{bibliography.bib}

\end{document}